\documentclass[letterpaper,journal]{IEEEtran}

\usepackage{amsmath,amssymb,amsfonts}
\usepackage{algorithm}
\usepackage{algorithmic}
\usepackage{graphicx}
\usepackage{textcomp}
\usepackage{xcolor}
\usepackage{booktabs}
\usepackage{multirow}
\usepackage{bm}
\usepackage{url}
\usepackage{cite}
\usepackage{hyperref}
\usepackage{stfloats}
\usepackage{balance}
\usepackage{placeins}

\begin{document}

\title{SARES-DEIM: Sparse Mixture-of-Experts Meets DETR for Clutter-Aware SAR Ship Detection}

\author{Fenghao~Song, Shaojing~Yang, and Xi~Zhou%
\thanks{Fenghao Song, Shaojing Yang, and Xi Zhou are with Yunnan Normal University, No. 768, Jucheng Avenue, Chenggong District, Kunming 650500, China.}%
\thanks{Corresponding author: Fenghao Song (e-mail: 2424420029@ynnu.edu.cn).}%
}

\markboth{Accepted for Publication in IEEE Journal of Selected Topics in Applied Earth Observations and Remote Sensing}%
{Song \MakeLowercase{\textit{et al.}}: SARES-DEIM for Clutter-Aware SAR Ship Detection}

\maketitle

\begin{abstract}
Ship detection in Synthetic Aperture Radar (SAR) imagery is challenged by coherent speckle noise, complex coastal clutter, and the prevalence of small-scale targets. Conventional detectors, primarily designed for optical imagery, often exhibit limited adaptability to SAR-specific degradation and suffer from the loss of fine-grained ship signatures during spatial downsampling. To address these limitations, we propose SARES-DEIM, a domain-aware detection framework grounded in the DEtection TRansformer (DETR) paradigm. Central to our approach is SARESMoE (SAR-aware Expert Selection Mixture-of-Experts), a module leveraging a sparse gating mechanism to selectively route features toward specialized wavelet, spatial, and frequency-oriented experts. This sparsely activated architecture enhances clutter-aware representation while keeping the active inference cost controlled. Furthermore, we introduce the Space-to-Depth Enhancement Pyramid (SDEP) neck to preserve high-resolution spatial cues from shallow stages and improve the localization of small targets. Experiments on HRSID and SAR-Ship-Dataset show that SARES-DEIM achieves competitive results across both public benchmarks, with more prominent advantages on the challenging HRSID dataset. Specifically, SARES-DEIM achieves 76.4\% $mAP_{50:95}$ and 93.8\% $mAP_{50}$ on HRSID, while keeping the active computation at 72.6 GFLOPs and 136.5 FPS on a single NVIDIA A40 GPU with batch size 8; the batch-size-1 latency is 37.6 ms.
\end{abstract}


\section{Introduction}\label{sec:intro}

Synthetic Aperture Radar (SAR) ship detection~\cite{11271640,11258897,10468641,11016180,10811768,SUN2026105127,10964367,rs13214209} has become an important component of maritime situational awareness, providing all-weather and round-the-clock monitoring capabilities. Despite these advantages, the unique side-looking imaging mechanism of SAR introduces a key difficulty: \textit{the entanglement of target backscattering signatures with non-stationary background interference}. Specifically, pervasive coherent speckle noise, complex near-shore clutter, and extreme scale variations manifest as high false-alarm rates and missed detections, particularly for small-scale targets embedded in high-clutter coastal environments.

Existing SAR ship detectors typically follow two technical paradigms: the You Only Look Once (YOLO) series and DETR-style detectors. The YOLO series~\cite{yolov8,yolov11,yolov12}, while establishing popular benchmarks for one-stage architectures, usually depends on dense prediction grids, task-specific label assignment, and post-processing such as Non-Maximum Suppression (NMS). These components may generalize less effectively across diverse SAR sensors and sea conditions where scattering patterns vary significantly. Conversely, DETR-style detectors~\cite{carion2020detr,zhu2020deformabledetr,lv2023detrs,zhang2023dino,huang2025deim,peng2025dfine} offer a streamlined end-to-end framework, eliminating the need for complex hand-crafted priors by formulating detection as a bipartite matching problem. However, most general-purpose DETR variants are primarily designed for optical imagery and do not explicitly incorporate SAR-specific scattering characteristics. Consequently, static weight-sharing operators may struggle to distinguish fine-grained ship signatures from low-Signal-to-Noise Ratio (SNR) backgrounds, which can limit detection performance in complex maritime surveillance applications.

In this work, we argue that the primary challenge in SAR detection lies in the \textit{diverse representation requirements} across heterogeneous maritime scenes. Different maritime backgrounds can induce distinct routing preferences: cluttered near-shore scenes may require stronger shallow-stage denoising to suppress coastal interference, whereas relatively homogeneous offshore scenes may benefit more from preserving compact target structures at shallow levels. Recent research underscores the critical role of frequency-domain interactions and coordinated multi-frequency modules in enhancing object representations for both aerial and SAR imagery~\cite{qiao2022novel,weng2023novel,weng2024enhancing}. Motivated by these insights, we adopt a divide-and-conquer strategy via the Mixture-of-Experts (MoE) philosophy~\cite{jacobs1991moe,shazeer2017moe,lin2026yolomaster}. However, vanilla MoE architectures lack domain-specific guidance, leading to suboptimal expert utilization. To bridge this gap, we propose SARES-DEIM, featuring a SAR-aware Expert Selection MoE (SARESMoE) module. By incorporating specialized wavelet, spatial, and frequency-domain experts governed by a sparse routing mechanism, SARESMoE adaptively decomposes features based on pooled image-level and pyramid-level spatial-frequency context. This sparse activation keeps the per-sample computation controlled while improving SAR-oriented feature refinement.

Furthermore, we identify that downsampling in standard backbones may discard fine-grained cues that are important for localizing small-scale ships. While recent efforts have explored location-refined feature pyramids to mitigate localization degradation in remote sensing~\cite{li2024lr}, we specifically design the Space-to-Depth Enhancement Pyramid (SDEP) neck to reduce the detail loss associated with downsampling in DETR-style architectures. Unlike traditional feature pyramids that primarily aggregate features from $1/8$ scale (P3) onwards, SDEP explicitly harvests high-resolution information from the P2 layer ($1/4$ scale). By employing Space-to-Depth Convolution (SPDConv)~\cite{sahi2022spdconv}, SDEP rearranges shallow spatial cues into the channel dimension and injects them into the detection head. This design is intended to help preserve the structural cues of small targets and reduce the information loss caused by strided convolutions.

While our framework incorporates established operations such as wavelet convolutions and space-to-depth transformations, our main contribution lies in two architectural designs tailored to SAR imagery: (1) a SAR-aware dual-branch routing mechanism that dynamically allocates these operations according to pooled image-level and pyramid-level spatial-frequency context, which is not directly provided by standard MoE networks; and (2) the SDEP neck, which introduces P2-level fine-grained spatial cues into the DETR-style pyramid to reduce semantic aliasing of small ships.

\noindent\textbf{Contributions.}
\begin{itemize}
    \item We present SARES-DEIM, a SAR-oriented DETR-based framework that bridges general-purpose object detection and SAR-domain requirements.
    \item We propose SARESMoE, a SAR-aware dual-branch mixture-of-experts module that routes features to heterogeneous experts according to pooled image-level and pyramid-level spatial-frequency context.
    \item We design the SDEP neck to introduce P2-level fine-grained spatial cues into a DETR-style pyramid for small-ship localization.
    \item Experiments, efficiency analysis, scale-wise statistics, and scene-wise analyses characterize the supervised in-domain accuracy gains, sparse-computation behavior, and limitations of SARES-DEIM.
\end{itemize}

\section{Related Work}

\subsection{YOLO-Series vs. DETR-Style Paradigms in SAR Detection}

SAR ship detection has rapidly evolved from heuristic-heavy frameworks to end-to-end prediction paradigms. The YOLO series~\cite{yolov8,yolov11,yolov12}, as representative one-stage detectors, has been widely adopted in maritime surveillance. Because YOLOv11 and YOLOv12 are recent releases, we treat them as highly competitive modern baselines rather than mature foundational references. These methods typically rely on dense prediction priors, task-specific assignment strategies, and Non-Maximum Suppression (NMS) to parse multiple detections. Although effective in certain scenarios, their robustness may decrease across heterogeneous SAR sensors and dense ship clusters with overlapping backscattering signatures.

Recently, DEtection TRansformer (DETR)~\cite{carion2020detr} and its variants have introduced a paradigm shift by eliminating anchors and NMS, instead formulating detection as a bipartite matching problem. Recent frameworks such as DEIM~\cite{huang2025deim} and D-FINE~\cite{peng2025dfine} develop dense one-to-one matching mechanisms and distribution-based localization refinement, making them competitive baselines for high-precision object detection. Unlike these general-purpose DETR variants, which are primarily designed for optical imagery, our \textsc{SARES-DEIM} retains their matching and localization components while adding SAR-oriented architectural inductive biases to improve feature representation under high clutter and low-SNR conditions.

\subsection{Feature Fusion and Conditional Computation}

Effective multi-scale representation and adaptive modeling are crucial for handling large scale variations of ships and the highly non-homogeneous nature of SAR backgrounds. Image and feature fusion have been broadly studied in remote sensing~\cite{albanwan2024imagefusionremotesensing}. For detection backbones, conventional feature pyramid structures, such as the Feature Pyramid Network (FPN)~\cite{lin2017fpn}, aggregate multi-scale semantics effectively, but often suffer from structural information loss caused by aggressive strided downsampling. To alleviate the degradation of fine-grained details, methods such as SPDConv~\cite{sahi2022spdconv} employ a space-to-depth transformation to rearrange spatial samples into channels without directly discarding them during the rearrangement.

Meanwhile, adaptive conditional computation has shown useful behavior across a broad range of vision tasks, including controllable image editing, layout-consistent generation, and fine-grained garment synthesis~\cite{shen2025imagedit,shen2025imagharmony,shen2025imaggarment,shen2024advancing,shen2024imagpose,shen2025imagdressing,shenlong}. These advances suggest that condition-aware feature transformation can improve representational flexibility when the input distribution is complex or highly diverse. Inspired by this line of research, the Mixture-of-Experts (MoE)~\cite{jacobs1991moe,shazeer2017moe,lin2026yolomaster} paradigm provides a natural mechanism for expanding model capacity through sparse and adaptive routing. However, conventional MoE gating usually depends solely on spatial features, which may overlook the distinctive frequency-domain backscattering characteristics of SAR targets, thereby leading to suboptimal expert selection in challenging maritime scenes.

To achieve multi-scale detection and scene-aware representation, \textsc{SARES-DEIM} introduces two key designs. First, the SDEP neck leverages the space-to-depth mechanism to explicitly exploit fine-grained cues from the high-resolution P2 layer, helping preserve structural signatures of small-scale ships during feature aggregation. Second, the SARESMoE module adopts a SAR-aware expert selection strategy to adaptively model diverse scattering patterns with specialized wavelet, spatial, frequency, and hybrid experts. By combining detail-preserving fusion with domain-specific dynamic routing, our method seeks a better balance between SAR-aware representation design and practical detection performance.
\section{Proposed Method}

\subsection{Overview}

Given an input SAR image $\mathbf{I}\in\mathbb{R}^{H\times W\times C}$, our detector follows an end-to-end set prediction paradigm. A backbone network extracts multi-scale feature maps $\{\mathbf{F}_2, \mathbf{F}_3, \mathbf{F}_4, \mathbf{F}_5\}$, where $\mathbf{F}_\ell \in \mathbb{R}^{C_\ell \times H_\ell \times W_\ell}$ denotes the features at pyramid level $\ell$ with a spatial stride $s_\ell = 2^\ell$.

\begin{figure*}[t]
  \centering
  \includegraphics[width=\linewidth]{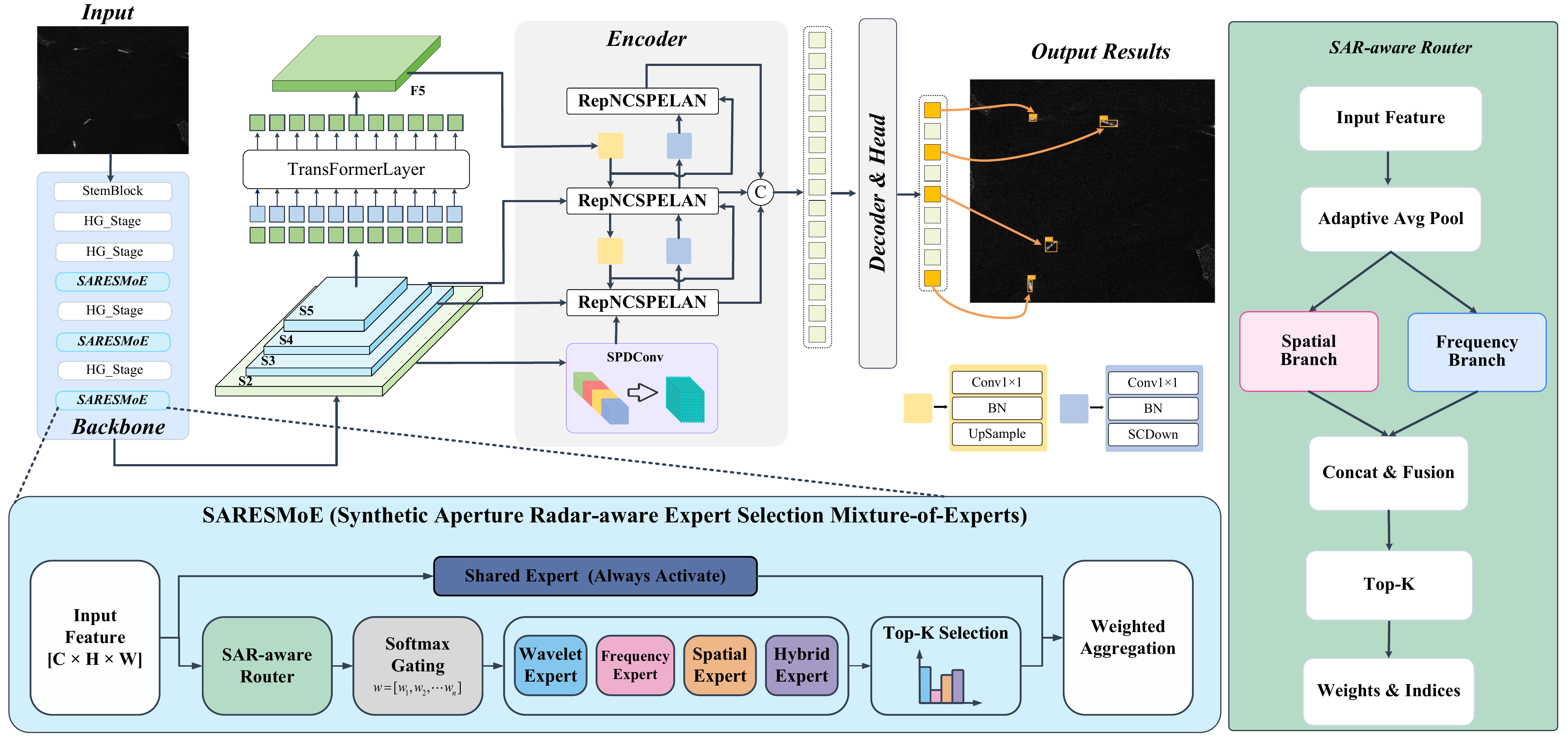}
\caption{\textsc{SARES-DEIM} overview. The backbone produces the
           multi-scale features, SARESMoE performs image-level Top-$k$
           expert selection with an always-active Shared Expert, and SDEP
           injects P2 detail into the P3 pathway before the encoder and
           DETR-style decoder/head.}
  \label{fig:overview}
\end{figure*}

To adapt the model to the unique backscattering characteristics of SAR imagery, such as coherent speckle noise and coastal clutter, we integrate the SAR-aware Expert Selection MoE (SARESMoE) module into selected backbone pyramid stages. This module produces enhanced features $\tilde{\mathbf{F}}_\ell = \mathrm{SARESMoE}(\mathbf{F}_\ell)$, enabling sample-adaptive representation learning by expanding the model's stored expert capacity for non-homogeneous backgrounds. Importantly, the router outputs one expert distribution per input image at each pyramid level, so all spatial locations at the same level share the same selected sparse experts and routing weights. Subsequently, the Space-to-Depth Enhancement Pyramid (SDEP) neck aggregates these multi-scale features. Notably, SDEP explicitly injects high-resolution, fine-grained cues from $\mathbf{F}_2$ into the detection pyramid to improve the localization of diminutive ships, yielding a refined feature set $\{\mathbf{P}_3, \mathbf{P}_4, \mathbf{P}_5\}$ with a unified channel dimension $d$. As illustrated in Fig.~\ref{fig:overview}, a DETR-style decoder~\cite{carion2020detr} finally processes $N$ object queries against these pyramid features to produce the prediction set $\hat{\mathcal{Y}}=\{(\hat{\mathbf{s}}_i, \hat{\mathbf{b}}_i)\}_{i=1}^{N}$, where $\hat{\mathbf{s}}_i \in \mathbb{R}^{K}$ and $\hat{\mathbf{b}}_i \in [0, 1]^4$ denote the classification logits and normalized bounding boxes, respectively. Here, $K$ is the number of foreground classes ($K=1$ for the ship-only experiments), $N$ is the fixed number of decoder queries, and $d$ is the unified channel dimension of the decoder input. Unless otherwise specified, the DEIM-S~\cite{huang2025deim} detection head, object-query decoder, matching strategy, and base detection losses are kept unchanged; our modifications are confined to the SARESMoE/SDEP feature pathway and the MoE auxiliary terms described below.

\subsection{SARESMoE: SAR-aware Expert Selection Mixture-of-Experts}

Standard convolutional layers apply static filters across the entire image, ignoring the non-stationary nature of SAR signals. In maritime scenarios, targets and clutter exhibit different spectral and spatial characteristics across different scales. To address this, we propose the SARESMoE module, which organizes established wavelet-domain, spatial, and frequency-domain operators within a SAR-aware, scale-aware mixture-of-experts framework. The operators themselves are not claimed as new; the novelty lies in their SAR-oriented expert organization and dual-branch routing.

Formally, let $\mathbf{X} \in \mathbb{R}^{C \times H \times W}$ be the input feature map for one image, with the batch dimension omitted for clarity. SARESMoE comprises an always-active Shared Expert that maintains a consistent semantic representation and a bank of sparse experts managed by a SAR-aware router.

\noindent\textbf{SAR-Aware Routing Mechanism.}
The router follows a dual-branch design that jointly analyzes frequency-domain and spatial-domain cues before producing sparse expert weights. To reduce routing overhead, the input feature is first adaptively pooled to a small spatial grid. The frequency branch computes the log-amplitude spectrum of the pooled feature and refines it with a lightweight convolutional analyzer, while the spatial branch directly applies a convolutional analyzer to the pooled feature. The two branches are then concatenated and fused before a $1\times1$ routing head produces expert logits:

\begin{align}
\mathbf{X}_p &= \mathcal{P}_{s}(\mathbf{X}), \nonumber\\
\mathbf{U}_{f} &= \phi_f\!\left(\log\left(|\mathcal{F}_{2D}(\mathbf{X}_p)|+\varepsilon\right)\right), \nonumber\\
\mathbf{U}_{s} &= \phi_s(\mathbf{X}_p), \nonumber\\
\mathbf{G} &= \phi_g\!\left([\mathbf{U}_{f};\mathbf{U}_{s}]\right), \nonumber\\
\mathbf{z} &= \mathrm{GAP}\!\left(\psi_r(\mathbf{G})\right), \quad
\boldsymbol{\pi} = \mathrm{Softmax}(\mathbf{z}),
\label{eq:router}
\end{align}

where $\mathcal{P}_{s}$ denotes adaptive average pooling to an $s \times s$ grid ($s=4$ in our implementation), $\mathcal{F}_{2D}$ is the two-dimensional Fourier transform, $\phi_f$ and $\phi_s$ denote the frequency and spatial analyzers, $\phi_g$ denotes the branch-fusion layer, $\psi_r$ is the routing head, GAP denotes global average pooling, and $[\cdot;\cdot]$ denotes channel-wise concatenation. If $N_e$ denotes the number of sparse experts at the current pyramid level, then $\mathbf{z},\boldsymbol{\pi}\in\mathbb{R}^{N_e}$. Concretely, the router uses $C_r=\max(\lfloor C/8\rfloor,8)$ hidden channels. Both $\phi_f$ and $\phi_s$ are implemented as convolution--batch-normalization--Sigmoid Linear Unit (Conv-BN-SiLU) analyzers with $3\times3$ kernels; $\phi_g$ is a $1\times1$ Conv-BN-SiLU fusion layer; and $\psi_r$ is a $1\times1$ Conv-BN routing head whose convolution is initialized from $\mathcal{N}(0,0.01^2)$. During training, zero-mean Gaussian perturbation with standard deviation 1.0 is added to the logits before Softmax; this perturbation is disabled during inference. Thus, (\ref{eq:router}) produces one image-level distribution for each pyramid level. The routing logits and weights are then sparsified and renormalized as

\begin{equation}
\bar{\pi}_e =
\frac{\pi_e}{\sum_{j \in \mathcal{T}_k(\boldsymbol{\pi})}\pi_j+\varepsilon},
\quad e \in \mathcal{T}_k(\boldsymbol{\pi}),
\qquad \bar{\pi}_e=0\quad e\notin\mathcal{T}_k(\boldsymbol{\pi}),
\label{eq:topk}
\end{equation}

where $\mathcal{T}_k(\boldsymbol{\pi})$ is the set of $k$ selected expert indices for the current input and $\varepsilon=10^{-6}$. Equation~(\ref{eq:topk}) leaves unselected experts with zero weight. The final output combines the shared path and the selected sparse expert paths:
\begin{equation}
\mathbf{Y} = \mathrm{S}_{\mathrm{shared}}(\mathbf{X}) + \sum_{e \in \mathcal{T}_k(\boldsymbol{\pi})} \bar{\pi}_e \cdot \mathrm{E}_e(\mathbf{X}).
\label{eq:sarmoe_agg}
\end{equation}
Equation~(\ref{eq:sarmoe_agg}) combines the always-active Shared Expert path with the selected sparse paths; the Shared Expert is not included in the Top-$k$ competition. During training, the original DEIM detection objective is retained. The complete objective is

\begin{equation}
\begin{aligned}
\mathcal{L}_{\mathrm{total}} &= \mathcal{L}_{\mathrm{DEIM}}+\lambda_{\mathrm{bal}}\mathcal{L}_{\mathrm{bal}}+\lambda_z\mathcal{L}_z,\\
\lambda_{\mathrm{bal}} &= 0.01,\qquad \lambda_z=10^{-3}.
\end{aligned}
\label{eq:loss}
\end{equation}

Here, $\mathcal{L}_{\mathrm{bal}}$ and $\mathcal{L}_z$ are computed from the router probabilities, logits, and selected indices at each pyramid level. Equation~(\ref{eq:loss}) augments the unchanged DEIM detection objective with the MoE load-balancing and router $z$-loss terms. In this implementation, the Shared Expert $\mathrm{S}_{\mathrm{shared}}$ uses a $3\times3$ Conv-BN-SiLU block followed by a lightweight channel-attention projection inspired by Frequency-Spatial Attention (FSA)~\cite{fsanet2025}. The Wavelet Expert uses a one-level Haar WTConv with a $5\times5$ kernel, the Frequency Expert uses Frequency-Adaptive Dilated Convolution (FADC)-based frequency selection with low-frequency attention, the Spatial Expert uses a GhostExpert-style spatial operator, and the Hybrid Expert concatenates frequency-selection and Haar-wavelet branches before $1\times1$ fusion. Unless otherwise specified, \textsc{SARES-DEIM} uses a progressive expert-bank configuration of 4/8/16 sparse experts at P3/P4/P5 with Top-2 activation. The Shared Expert is always active, while the sparse expert bank is selectively activated by the router. During inference, after obtaining the Top-$k$ expert indices, only the selected sparse experts are executed for the corresponding batch samples; unselected sparse experts are skipped rather than evaluated and masked afterward. The selected outputs are weighted by the normalized routing weights and accumulated into the output tensor. Therefore, Params report the stored expert-bank capacity, whereas FLOPs, FPS, and batch-size-1 latency report the actually executed active inference path under sparse routing.

\noindent\textbf{Scale-Aware Expert Selection.}
We design heterogeneous experts by leveraging established dynamic operators and assign them via a \emph{Scale-Aware Allocation Strategy}:
\begin{itemize}
\item Wavelet \& Spatial Experts (Assigned to P3): P3 is the highest-resolution pyramid level processed by SARESMoE, so it retains more small-ship signatures but is also more sensitive to speckle and coastal clutter. We employ WTConv~\cite{wtconv2024} as the \emph{Wavelet Expert} to support wavelet-domain denoising and high-frequency interference separation, and utilize GhostNet modules~\cite{ghostnet2020} as \emph{Spatial Experts} to preserve local shape and compact backscattering structures. The P3 expert bank alternates these two expert types, allowing the router to balance clutter suppression and structure preservation at the shallow feature level.
\item Frequency \& Hybrid Experts (Assigned to P4/P5): P4 and P5 are lower-resolution and semantically richer pyramid levels, where broader spectral responses and multi-scale scattering patterns become more relevant. We adopt the frequency selection mechanism from FADC~\cite{fadc2024} to construct \emph{Frequency Experts}, which performs multi-scale frequency selection filtering to adaptively suppress spectral clutter. Additionally, \emph{Hybrid Experts} combine parallel frequency-domain and wavelet-domain branches to handle complex multi-scale scattering scenarios. The P4 bank alternates Hybrid and Frequency Experts, whereas the P5 bank alternates Frequency and Hybrid Experts.
\end{itemize}
This design encourages the model to select between detail-preserving and clutter-suppression-oriented experts according to the feature level and input context, and we treat this as a SAR-aware design rather than as a strict proof of physical interpretability.

\subsection{SDEP: Space-to-Depth Enhancement Pyramid}

Detecting diminutive ships in SAR imagery is hindered by the aliasing induced by standard strided convolutions, which may erode high-frequency scattering highlights. To address this, we propose the SDEP neck, which uses a space-to-depth detail transmission mechanism to explicitly inject fine-grained spatial cues from shallow layers into the semantic pyramid.

\noindent\textbf{Space-to-Depth Transformation.}
The core of SDEP is the Space-to-Depth (SPD) operation~\cite{sahi2022spdconv}, which rearranges spatial pixels into the channel dimension without discarding spatial samples during the rearrangement. Given the high-resolution backbone feature $\mathbf{F}_2 \in \mathbb{R}^{C_2 \times H_2 \times W_2}$ at stride 4, the SPD transformation in (\ref{eq:spd}) reshapes it as:

\begin{equation}
\begin{aligned}
\mathrm{SPD}(\mathbf{F}_2)
  &= \mathrm{Rearrange}_{2\times2}(\mathbf{F}_2) \\
  &\in \mathbb{R}^{4C_2 \times \frac{H_2}{2} \times \frac{W_2}{2}},
\end{aligned}
\label{eq:spd}
\end{equation}

where $\mathrm{Rearrange}_{2\times2}$ denotes the standard space-to-depth rearrangement: every $2 \times 2$ spatial neighborhood is stacked along the channel axis. This operation halves the spatial resolution to match the P3 stride while preserving the spatial samples through channel rearrangement, avoiding the lossy pooling or strided convolution that could discard fine-grained scattering signatures.

\noindent\textbf{Detail-Preserving Fusion.}
After the SPD transformation, SDEP uses the convolutional projection inside SPDConv to compress the expanded P2 channels to the pre-fusion P3-channel dimension $C_3$ and then concatenates the resulting detail feature with the SARESMoE-enhanced P3 feature. We denote this immediate P2-to-P3 fusion result as $\mathbf{P}_3^{\mathrm{sdep}}$ to distinguish it from the final decoder input $\mathbf{P}_3$; the complete fusion is defined in (\ref{eq:sdep_fuse}):

\begin{equation}
\begin{aligned}
\mathbf{D}_2 &= \mathrm{SPDConv}(\mathbf{F}_2)
= \mathrm{Conv}_{3\times3}\!\left(\mathrm{SPD}(\mathbf{F}_2)\right),\\
\mathbf{D}_2 &\in \mathbb{R}^{C_3 \times H_3 \times W_3},\\
\mathbf{Z}_3 &= [\mathbf{D}_2;\tilde{\mathbf{F}}_3],\\
\mathbf{P}_3^{\mathrm{sdep}} &= \Phi_3^{\mathrm{sdep}}(\mathbf{Z}_3),\\
\mathbf{P}_3^{\mathrm{sdep}} &\in \mathbb{R}^{d \times H_3 \times W_3},
\end{aligned}
\label{eq:sdep_fuse}
\end{equation}

where $\tilde{\mathbf{F}}_3 \in \mathbb{R}^{C_3 \times H_3 \times W_3}$ denotes the SARESMoE-enhanced P3 feature, $[\cdot;\cdot]$ denotes channel-wise concatenation, and $\Phi_3^{\mathrm{sdep}}$ is the local neck refinement block that maps the concatenated feature to the unified dimension $d$. In our implementation, this corresponds to the SPDConv block from P2, i.e., space-to-depth rearrangement followed by a $3\times3$ convolutional projection, concatenation with the P3 feature, and a RepNCSPELAN4 refinement block inherited from the DEIM-S neck. Here, $H_3=H_2/2$ and $W_3=W_2/2$ under the stride-4/stride-8 setting. The intermediate $\mathbf{P}_3^{\mathrm{sdep}}$ then participates in the subsequent top-down and bottom-up fusion pathway, whose final outputs $\{\mathbf{P}_3, \mathbf{P}_4, \mathbf{P}_5\}$ are consumed by the DETR-style decoder. By explicitly bridging the P2 and P3 layers through this space-to-depth mechanism, SDEP helps preserve local structural cues of diminutive targets throughout the feature aggregation stages.

\begin{table*}[t]
  \caption{Detection performance and efficiency comparison on HRSID~\cite{wei2020hrsid}.
           Bold indicates the best detection metric among listed entries. Baseline entries marked with $^\ast$ are locally reproduced under our implementation setting; unmarked entries are reported from the corresponding cited papers when available. SARES-DEIM is the proposed method evaluated under the same setting. Base denotes the direct DEIM-S~\cite{huang2025deim} architectural baseline.}
  \label{tab:hrsid}
  \centering
  \footnotesize
  \renewcommand{\arraystretch}{1.15}
  \setlength{\tabcolsep}{2.4pt}
  \begin{tabular*}{\linewidth}{@{\extracolsep{\fill}}lccccccc@{}}
    \toprule
    Model & $mAP_{50:95}$\,(\%) & $mAP_{50}$\,(\%) & $P$\,(\%) & $R$\,(\%) & Params (M) & FLOPs (G) & FPS \\
    \midrule
    Faster R-CNN~\cite{FasterRCNN}   & -- & 88.0 & 82.1 & -- & 136.7 & 147.8 & -- \\
    SSD~\cite{SSD}                  & -- & 90.1 & 81.0 & -- & 32.0 & 70.0 & -- \\
    FCOS~\cite{FCOS}                & -- & 81.1 & 83.6 & -- & 36.3 & 128.0 & -- \\
    CSCF-Net~\cite{CSCF-Net}        & -- & 91.5 & 90.7 & -- & 2.61 & 9.4 & -- \\
    SAR-D-FINE~\cite{SAR-D-FINE}    & 72.0 & 90.8 & 91.2 & -- & 11.8 & 31.0 & 109.0 \\
    OptiSAR-Net~\cite{OptiSAR-Net}  & 65.4 & 90.2 & -- & -- & 2.7 & 11.8 & 90.6 \\
    FINet~\cite{FINet}              & -- & 90.5 & 68.6 & 72.0 & -- & -- & -- \\
    CoAM-Darknet~\cite{CoAM-Darknet} & -- & 92.7 & 92.7 & \textbf{88.1} & 65.8 & 123.5 & -- \\
    \midrule
    YOLOv8-S$^\ast$~\cite{yolov8_ultralytics}  & 61.7 & 88.2 & 90.2 & 78.3 & 9.8 & 23.6 & 212.8 \\
    YOLOv8-M$^\ast$~\cite{yolov8_ultralytics}  & 62.7 & 88.1 & 87.6 & 80.4 & 23.2 & 67.8 & 204.8 \\
    YOLOv11-S$^\ast$~\cite{yolo11_ultralytics} & 61.6 & 87.9 & 89.6 & 78.2 & 9.4 & 21.5 & 243.9 \\
    YOLOv11-M$^\ast$~\cite{yolo11_ultralytics} & 63.2 & 88.1 & 89.3 & 78.6 & 20.0 & 68.2 & 172.4 \\
    RT-DETR-L$^\ast$~\cite{lv2023detrs}        & 59.1 & 87.1 & 87.1 & 79.1 & 28.8 & 104.8 & 106.4 \\
    D-FINE-S$^\ast$~\cite{peng2025dfine}       & 70.2 & 88.2 & 92.5 & 82.7 & 10.2 & 24.8 & 227.4 \\
    D-FINE-M$^\ast$~\cite{peng2025dfine}       & 75.1 & 91.0 & 91.6 & 86.6 & 19.2 & 56.3 & 143.8 \\
    DEIM-S (Base)$^\ast$~\cite{huang2025deim}  & 73.0 & 92.0 & 92.4 & 84.4 & 10.2 & 24.8 & 237.9 \\
    DEIM-M$^\ast$~\cite{huang2025deim}         & 75.6 & 93.1 & 92.8 & 87.8 & 19.2 & 56.3 & 143.4 \\
    \midrule
    \textbf{SARES-DEIM}                & \textbf{76.4} & \textbf{93.8} & \textbf{93.1} & 88.0 & 90.8 & 72.6 & 136.5 \\
    \bottomrule
  \end{tabular*}
\end{table*}

\section{Experiments and Analysis}

To evaluate the effectiveness of the proposed \textsc{SARES-DEIM}, we conduct experiments on two widely-used SAR ship detection benchmarks, namely \emph{HRSID}~\cite{wei2020hrsid} and \emph{SAR-Ship-Dataset}~\cite{wang2019sar}. Our method is compared against representative classical detectors, SAR-specific detectors, and recent YOLO-series and DETR-style baselines following the protocol described below.

\noindent\textbf{Datasets.}
\emph{High-Resolution SAR Images Dataset (HRSID)}~\cite{wei2020hrsid} is a high-resolution SAR ship detection dataset containing 5,604 image patches and 16,951 ship instances. Cropped from panoramic Sentinel-1 and TerraSAR-X imagery, it features spatial resolutions from 1\,m to 5\,m, providing diverse scenarios including near-shore and far-sea environments for multi-scale evaluation.
\emph{SAR-Ship-Dataset}~\cite{wang2019sar} comprises 43,819 ship chips extracted from 102 Gaofen-3 and 108 Sentinel-1 images. With resolutions spanning from 3\,m to 25\,m and various polarization modes, it serves as a useful benchmark for evaluating adaptive representation capability across multiple sensors.
The SAR Ship Detection Dataset (SSDD)~\cite{zhang2021ssdd} is used only as an additional target dataset in the source-only cross-dataset diagnostic and is not used for training, validation, or model selection.

\noindent\textbf{Metrics.}
We employ Precision ($P$), Recall ($R$), and Average Precision (AP) as primary metrics, where true positives (TP), false positives (FP), and false negatives (FN) give $P=\mathrm{TP}/(\mathrm{TP}+\mathrm{FP})$ and $R=\mathrm{TP}/(\mathrm{TP}+\mathrm{FN})$. AP is computed using the official Common Objects in Context (COCO) evaluation~\cite{lin2014microsoft} with at most 100 detections per image. We report $mAP_{50}$ at an Intersection over Union (IoU) threshold of 0.50 and $mAP_{50:95}$ averaged over IoU thresholds from 0.50 to 0.95 in increments of 0.05. For scale-wise analysis, $AP_S$, $AP_M$, and $AP_L$ follow the standard COCO object-area ranges, where the bounding-box area $A=w\times h$ is measured in the evaluation annotation coordinate system: small objects satisfy $A<32^2$ pixels$^2$, medium objects satisfy $32^2\leq A<96^2$ pixels$^2$, and large objects satisfy $A\geq96^2$ pixels$^2$. These image-space bins correspond to square-side equivalents of below 32 pixels, 32 to below 96 pixels, and at least 96 pixels, respectively. They do not define physical vessel-length ranges because the ground sampling distance varies across sensors and datasets. Precision and Recall are additionally computed at a confidence threshold of 0.25 and an IoU threshold of 0.50 using score-descending greedy one-to-one matching, with at most 100 detections retained per image; no target-domain threshold tuning is used.

\noindent\textbf{Implementation Details.}
\textsc{SARES-DEIM} is implemented in PyTorch with CUDA 11.8. All locally reproduced models are trained on a single NVIDIA A40 GPU (48\,GB VRAM). We adopt the AdamW optimizer with a base learning rate of $8 \times 10^{-4}$ and a backbone learning rate of $4 \times 10^{-4}$. The network is trained for 300 epochs with a batch size of 8, and input images are resized to $640 \times 640$. Strong augmentations (Mosaic and Mixup) are utilized initially but are disabled after epoch 288 to ensure the model converges on the original data distribution and refines localization. Unless otherwise stated, all locally reproduced single-run experiments and ablations use random seed 0. For the routing-robustness study, we additionally use seeds 13, 21, and 42 under the same training and evaluation protocol.

To ensure a controlled comparison for locally reproduced baselines, all YOLO-series and DETR-series entries reported in our tables were trained and evaluated on the same hardware under a unified protocol. For these locally reproduced entries, we used the official dataset partitions and the same input resolution ($640 \times 640$), batch size (8), training epochs (300), and augmentation schedule whenever supported by the corresponding framework. Results and efficiency values for traditional and certain SAR-specific networks are not remeasured in our environment; instead, they are cited directly from the corresponding papers when available and left unmarked in Tables~\ref{tab:hrsid} and~\ref{tab:sar_ship}. Baseline entries marked with $^\ast$ are locally reproduced under our implementation setting. Therefore, cross-family comparisons involving literature-reported entries should be interpreted as reference comparisons, whereas the most controlled architectural comparisons are against the locally reproduced DEIM, D-FINE, RT-DETR, and YOLO baselines. Parameters (Params), floating-point operations (FLOPs), and frames per second (FPS) are reported using millions (M), giga operations (G), and frames per second, respectively. Params/FLOPs/FPS of the locally reproduced YOLO-series and DETR-series baselines are measured under the same input resolution ($640 \times 640$) and hardware setting (single NVIDIA A40 GPU, batch size 8). We additionally report batch-size-1 latency for these locally reproduced entries, computed as $1000/\mathrm{FPS}_{\mathrm{BS}=1}$ under the same input resolution after warm-up. Since \textsc{SARES-DEIM} is constructed by inserting SARESMoE and SDEP into the DEIM-S~\cite{huang2025deim} detection framework, DEIM-S~\cite{huang2025deim} is treated as the direct architectural baseline. DEIM-M and D-FINE-M are included as medium-scale reference models because their FLOPs are closer to \textsc{SARES-DEIM} than the S variants, providing same-tier active-computation context rather than serving as direct baselines. For SARESMoE, Params indicate the stored expert-bank capacity, while FLOPs, FPS, and latency are used as the main indicators of active inference cost. The reported 72.6 GFLOPs corresponds to the computational cost of the entire model under the default routing configuration, rather than an average over evaluated samples. The router's computation is included.

\begin{table*}[t]
  \caption{Detection performance and efficiency comparison on SAR-Ship-Dataset~\cite{wang2019sar}.
           Bold indicates the best detection metric among listed entries. Baseline entries marked with $^\ast$ are locally reproduced under our implementation setting; unmarked entries are reported from the corresponding cited papers when available. SARES-DEIM is the proposed method evaluated under the same setting. Base denotes the direct DEIM-S~\cite{huang2025deim} architectural baseline.}
  \label{tab:sar_ship}
  \centering
  \footnotesize
  \renewcommand{\arraystretch}{1.15}
  \setlength{\tabcolsep}{2.4pt}
  \begin{tabular*}{\linewidth}{@{\extracolsep{\fill}}lccccccc@{}}
    \toprule
    Model & $mAP_{50:95}$\,(\%) & $mAP_{50}$\,(\%) & $P$\,(\%) & $R$\,(\%) & Params (M) & FLOPs (G) & FPS \\
    \midrule
    Faster R-CNN~\cite{FasterRCNN}       & -- & 86.3 & 80.7 & -- & 136.7 & 147.8 & -- \\
    SSD~\cite{SSD}                      & -- & 88.9 & 87.1 & -- & 32.0 & 70.0 & -- \\
    FCOS~\cite{FCOS}                    & -- & 87.5 & 89.1 & -- & 36.3 & 128.0 & -- \\
    CoAM-Darknet~\cite{CoAM-Darknet}    & -- & 96.0 & 93.7 & 95.3 & 65.8 & 123.5 & -- \\
    \midrule
    YOLOv8-S$^\ast$~\cite{yolov8_ultralytics}  & 66.1 & 97.1 & 94.5 & 93.6 & 9.8 & 23.6 & 212.8 \\
    YOLOv8-M$^\ast$~\cite{yolov8_ultralytics}  & 66.2 & 95.5 & 91.5 & 90.8 & 23.2 & 67.8 & 204.8 \\
    YOLOv11-S$^\ast$~\cite{yolo11_ultralytics} & 66.4 & 97.0 & 94.4 & 93.8 & 9.4 & 21.5 & 243.9 \\
    YOLOv11-M$^\ast$~\cite{yolo11_ultralytics} & 61.5 & 87.8 & 91.6 & 90.8 & 20.0 & 68.2 & 172.4 \\
    RT-DETR-L$^\ast$~\cite{lv2023detrs}        & 67.3 & 96.8 & 95.2 & 94.9 & 28.8 & 104.8 & 106.4 \\
    D-FINE-S$^\ast$~\cite{peng2025dfine}       & 71.1 & 96.8 & 96.4 & 95.3 & 10.2 & 24.8 & 227.4 \\
    D-FINE-M$^\ast$~\cite{peng2025dfine}       & 71.2 & 97.0 & \textbf{96.7} & 95.5 & 19.2 & 56.3 & 143.8 \\
    DEIM-S (Base)$^\ast$~\cite{huang2025deim}  & 71.3 & 97.6 & 96.5 & 96.1 & 10.2 & 24.8 & 237.9 \\
    DEIM-M$^\ast$~\cite{huang2025deim}         & 71.6 & 97.4 & 96.6 & 96.1 & 19.2 & 56.3 & 143.4 \\
    \midrule
    \textbf{SARES-DEIM}                & \textbf{71.7} & \textbf{98.1} & \textbf{96.7} & \textbf{96.2} & 90.8 & 72.6 & 136.5 \\
    \bottomrule
  \end{tabular*}
\end{table*}

\subsection{Comparison with Representative Methods}\label{sec:sota_comparison}

\subsubsection{Comparisons on HRSID}

As summarized in Table~\ref{tab:hrsid}, our model is compared against classical architectures (Faster R-CNN~\cite{FasterRCNN}, SSD~\cite{SSD}, and FCOS~\cite{FCOS}), SAR-specific methods (CSCF-Net~\cite{CSCF-Net}, SAR-D-FINE~\cite{SAR-D-FINE}, OptiSAR-Net~\cite{OptiSAR-Net}, FINet~\cite{FINet}, and CoAM-Darknet~\cite{CoAM-Darknet}), general-purpose YOLO-series baselines (YOLOv8~\cite{yolov8_ultralytics} and YOLOv11~\cite{yolo11_ultralytics}), and DETR-style baselines (RT-DETR~\cite{lv2023detrs}, D-FINE~\cite{peng2025dfine}, and DEIM~\cite{huang2025deim}). DEIM-S~\cite{huang2025deim} is the direct architectural baseline of \textsc{SARES-DEIM}, while DEIM-M and D-FINE-M serve as medium-scale references for the same active-computation tier. Locally reproduced baseline entries are marked with $^\ast$ in the table, while unmarked entries are taken from the corresponding cited papers when available. The proposed SARES-DEIM row is not star-marked because it is our method evaluated under the same implementation setting. The table also reports Params, FLOPs, and batch-size-8 FPS to clarify the efficiency trade-off of the proposed MoE-based detector; batch-size-1 latency is reported separately in Table~\ref{tab:bs1_latency}.

Specifically, \textsc{SARES-DEIM} achieves $mAP_{50:95}$ of 76.4\%, $mAP_{50}$ of 93.8\%, Precision of 93.1\%, and Recall of 88.0\%. It obtains the highest $mAP_{50:95}$, $mAP_{50}$, and Precision among the listed entries with available detection metrics in Table~\ref{tab:hrsid}, while its Recall remains close to the highest reported value (88.0\% vs.\ 88.1\% for CoAM-Darknet). Against SAR-D-FINE, its $mAP_{50:95}$ is 4.4 percentage points higher (76.4\% vs.\ 72.0\%). Compared with the direct DEIM-S baseline, \textsc{SARES-DEIM} is 3.4 and 1.8 percentage points higher in $mAP_{50:95}$ and $mAP_{50}$, respectively. Compared with the medium-scale DEIM-M reference, it is 0.8 and 0.7 percentage points higher in the same metrics, with active computation of 72.6 GFLOPs versus 56.3 GFLOPs for DEIM-M. As further shown in Table~\ref{tab:topk_sparsity}, activating all stored experts would increase the cost to 126.0 GFLOPs. These results indicate that SARESMoE and SDEP improve the DEIM-S baseline, while sparse activation keeps the active computation below full expert activation.

\subsubsection{Comparisons on SAR-Ship-Dataset}

On the diverse SAR-Ship-Dataset~\cite{wang2019sar}, we compare with representative baselines following the same reporting convention: baseline entries marked with $^\ast$ are locally reproduced, while unmarked traditional and SAR-specific entries are cited from the literature when available. As shown in Table~\ref{tab:sar_ship}, \textsc{SARES-DEIM} achieves competitive performance on this multi-sensor dataset.

Compared with the direct DEIM-S baseline, \textsc{SARES-DEIM} improves $mAP_{50:95}$ from 71.3\% to 71.7\% and $mAP_{50}$ from 97.6\% to 98.1\%. Compared with the medium-scale DEIM-M reference, the improvement is smaller but remains positive in $mAP_{50:95}$ (71.7\% vs.\ 71.6\%) and $mAP_{50}$ (98.1\% vs.\ 97.4\%). We therefore treat SAR-Ship-Dataset as evidence of competitive performance under its supervised within-dataset protocol, rather than as proof of strong cross-domain robustness.

The reported Params/FLOPs/FPS also clarify the computational trade-off. SARES-DEIM stores a larger parameter bank (90.8 M) because multiple SAR-oriented experts are retained. However, sparse routing restricts the active computation to 72.6 GFLOPs at an input resolution of $640 \times 640$ and yields 136.5 FPS on a single NVIDIA A40 GPU with batch size 8. The separately reported batch-size-1 latency is 37.6 ms. Thus, the proposed model trades additional parameter storage for adaptive representation capacity. The gap between stored Params and active FLOPs is the expected behavior of sparse MoE: the expert bank expands the available representation capacity, whereas only a subset of experts is activated for each input.

To complement the batch-size-8 throughput reported in Tables~\ref{tab:hrsid} and~\ref{tab:sar_ship}, Table~\ref{tab:bs1_latency} further reports batch-size-1 inference latency for the locally reproduced baselines. Compact YOLO variants show lower single-image latency, while \textsc{SARES-DEIM} stores more parameters because of the heterogeneous expert banks. However, only the Top-$k$ sparse experts are activated during inference, and the measured batch-size-1 latency remains 37.6 ms on the same NVIDIA A40 platform. This result indicates that \textsc{SARES-DEIM} improves SAR detection accuracy with an explicit complexity-latency trade-off, rather than aiming to minimize single-image latency alone.

\begin{table*}[t]
  \caption{Efficiency comparison with batch-size-1 latency for locally reproduced models. FPS is measured with batch size 8, while latency is computed from batch-size-1 throughput under the same $640 \times 640$ input resolution on a single NVIDIA A40 GPU. Bold indicates the best efficiency value in each column; lower is better for Params, FLOPs, and latency, whereas higher is better for FPS.}
  \label{tab:bs1_latency}
  \centering
  \footnotesize
  \renewcommand{\arraystretch}{1.15}
  \setlength{\tabcolsep}{5pt}
  \begin{tabular*}{\linewidth}{@{\extracolsep{\fill}}lcccc@{}}
    \toprule
    Model & Params (M) & FLOPs (G) & FPS (batch size 8) & Latency (ms, batch size 1) \\
    \midrule
    YOLOv8-S~\cite{yolov8_ultralytics}  & 9.8  & 23.6  & 212.8 & 11.2 \\
    YOLOv8-M~\cite{yolov8_ultralytics}  & 23.2 & 67.8  & 204.8 & 11.5 \\
    YOLOv11-S~\cite{yolo11_ultralytics} & \textbf{9.4}  & \textbf{21.5}  & \textbf{243.9} & \textbf{10.1} \\
    YOLOv11-M~\cite{yolo11_ultralytics} & 20.0 & 68.2  & 172.4 & 13.0 \\
    RT-DETR-L~\cite{lv2023detrs}        & 28.8 & 104.8 & 106.4 & 66.2 \\
    D-FINE-S~\cite{peng2025dfine}       & 10.2 & 24.8  & 227.4 & 29.1 \\
    D-FINE-M~\cite{peng2025dfine}       & 19.2 & 56.3  & 143.8 & 34.5 \\
    DEIM-S (Base)~\cite{huang2025deim}  & 10.2 & 24.8  & 237.9 & 28.8 \\
    DEIM-M~\cite{huang2025deim}         & 19.2 & 56.3  & 143.4 & 34.8 \\
    \midrule
    \textbf{SARES-DEIM}                & 90.8 & 72.6  & 136.5 & 37.6 \\
    \bottomrule
  \end{tabular*}
\end{table*}

\subsubsection{Limitations Under Source-Only Cross-Dataset Evaluation}

SARES-DEIM follows a fully supervised detection paradigm and is optimized with source-domain annotations. Accordingly, we present cross-dataset inference only as a source-only diagnostic under severe domain shift, not as a zero-shot generalization or deployment claim. Specifically, we train DEIM-S~\cite{huang2025deim} and SARES-DEIM on the official HRSID~\cite{wei2020hrsid} training split and directly evaluate the unchanged checkpoints on SAR-Ship-Dataset~\cite{wang2019sar} and SSDD~\cite{zhang2021ssdd}. HRSID-trained checkpoints encode ship predictions as category ID 1, whereas both target datasets encode ship ground truth as category ID 0. We therefore explicitly map prediction category 1 to target category 0 before evaluation and audit the unused prediction category 0 separately. AP is computed with the official COCO evaluation~\cite{lin2014microsoft} using at most 100 detections per image. Precision, Recall, and TP/FP/FN counts use score-descending greedy one-to-one matching at an IoU threshold of 0.50, also after retaining at most 100 detections per image. No target-domain training samples, validation samples, threshold calibration, or hyperparameter selection are used. All images are evaluated at the same input resolution of $640 \times 640$. We compare only DEIM-S~\cite{huang2025deim} and SARES-DEIM to isolate the relative effect of SARESMoE and SDEP under the same source-only protocol.

\begin{table*}[t]
  \caption{Source-only cross-dataset evaluation of DEIM-S~\cite{huang2025deim} and SARES-DEIM with explicit prediction-to-target category mapping from HRSID~\cite{wei2020hrsid} to SAR-Ship-Dataset~\cite{wang2019sar} and SSDD~\cite{zhang2021ssdd}. AP uses the official COCO evaluation~\cite{lin2014microsoft}. $P$, $R$, and the error counts use a confidence threshold of 0.25 and an IoU threshold of 0.50. AP, $P$, and $R$ are reported in percent. Bold indicates the better result for each target dataset; lower is better for FP and FN.}
  \label{tab:cross_dataset}
  \centering
  \scriptsize
  \renewcommand{\arraystretch}{1.12}
  \setlength{\tabcolsep}{3.2pt}
  \begin{tabular*}{\linewidth}{@{\extracolsep{\fill}}llccccrrr@{}}
    \toprule
    Target Test Set & Method & $mAP_{50:95}$\,(\%) & $mAP_{50}$\,(\%) & $P$\,(\%) & $R$\,(\%) & TP & FP & FN \\
    \midrule
    SAR-Ship-Dataset~\cite{wang2019sar} & DEIM-S (Base)~\cite{huang2025deim} & 15.9 & 49.1 & 28.2 & 62.9 & 9,628 & 24,530 & 5,670 \\
    SAR-Ship-Dataset~\cite{wang2019sar} & \textbf{SARES-DEIM} & \textbf{17.2} & \textbf{54.7} & \textbf{34.6} & \textbf{67.0} & \textbf{10,248} & \textbf{19,352} & \textbf{5,050} \\
    \midrule
    SSDD~\cite{zhang2021ssdd} & DEIM-S (Base)~\cite{huang2025deim} & 34.3 & 70.3 & 47.4 & 75.2 & 401 & 445 & 132 \\
    SSDD~\cite{zhang2021ssdd} & \textbf{SARES-DEIM} & \textbf{35.2} & \textbf{75.1} & \textbf{51.1} & \textbf{80.1} & \textbf{427} & \textbf{408} & \textbf{106} \\
    \bottomrule
  \end{tabular*}
\end{table*}

As shown in Table~\ref{tab:cross_dataset}, at the shared confidence threshold of 0.25, SARES-DEIM improves both Precision and Recall on both target datasets. On SAR-Ship-Dataset, Precision increases from 28.2\% to 34.6\% and Recall from 62.9\% to 67.0\%, corresponding to 620 more true positives, 5,178 fewer false positives, and 620 fewer false negatives. On SSDD, Precision increases from 47.4\% to 51.1\% and Recall from 75.2\% to 80.1\%, corresponding to 26 more true positives, 37 fewer false positives, and 26 fewer false negatives.

To verify that this result is not specific to threshold 0.25, Table~\ref{tab:cross_dataset_thresholds} reports three shared confidence thresholds without target-domain calibration. SARES-DEIM improves both Precision and Recall at all three thresholds on both target datasets.

\begin{table*}[t]
  \caption{Shared-confidence-threshold analysis for DEIM-S~\cite{huang2025deim} and SARES-DEIM on SAR-Ship-Dataset~\cite{wang2019sar} and SSDD~\cite{zhang2021ssdd} at an IoU threshold of 0.50. All values are reported in percent; bold indicates the better result at each shared threshold.}
  \label{tab:cross_dataset_thresholds}
  \centering
  \footnotesize
  \renewcommand{\arraystretch}{1.12}
  \setlength{\tabcolsep}{5pt}
  \begin{tabular*}{\linewidth}{@{\extracolsep{\fill}}lccccc@{}}
    \toprule
    Target Test Set & Confidence & DEIM-S $P$ & DEIM-S $R$ & SARES-DEIM $P$ & SARES-DEIM $R$ \\
    \midrule
    \multirow{3}{*}{SAR-Ship-Dataset~\cite{wang2019sar}}
      & 0.10 & 7.4 & 69.7 & \textbf{13.0} & \textbf{72.6} \\
      & 0.25 & 28.2 & 62.9 & \textbf{34.6} & \textbf{67.0} \\
      & 0.50 & 55.7 & 55.1 & \textbf{60.1} & \textbf{59.6} \\
    \midrule
    \multirow{3}{*}{SSDD~\cite{zhang2021ssdd}}
      & 0.10 & 11.2 & 82.4 & \textbf{20.1} & \textbf{86.5} \\
      & 0.25 & 47.4 & 75.2 & \textbf{51.1} & \textbf{80.1} \\
      & 0.50 & 80.4 & 63.2 & \textbf{82.1} & \textbf{70.4} \\
    \bottomrule
  \end{tabular*}
\end{table*}

\begin{figure}[t]
  \centering
  \includegraphics[width=\columnwidth]{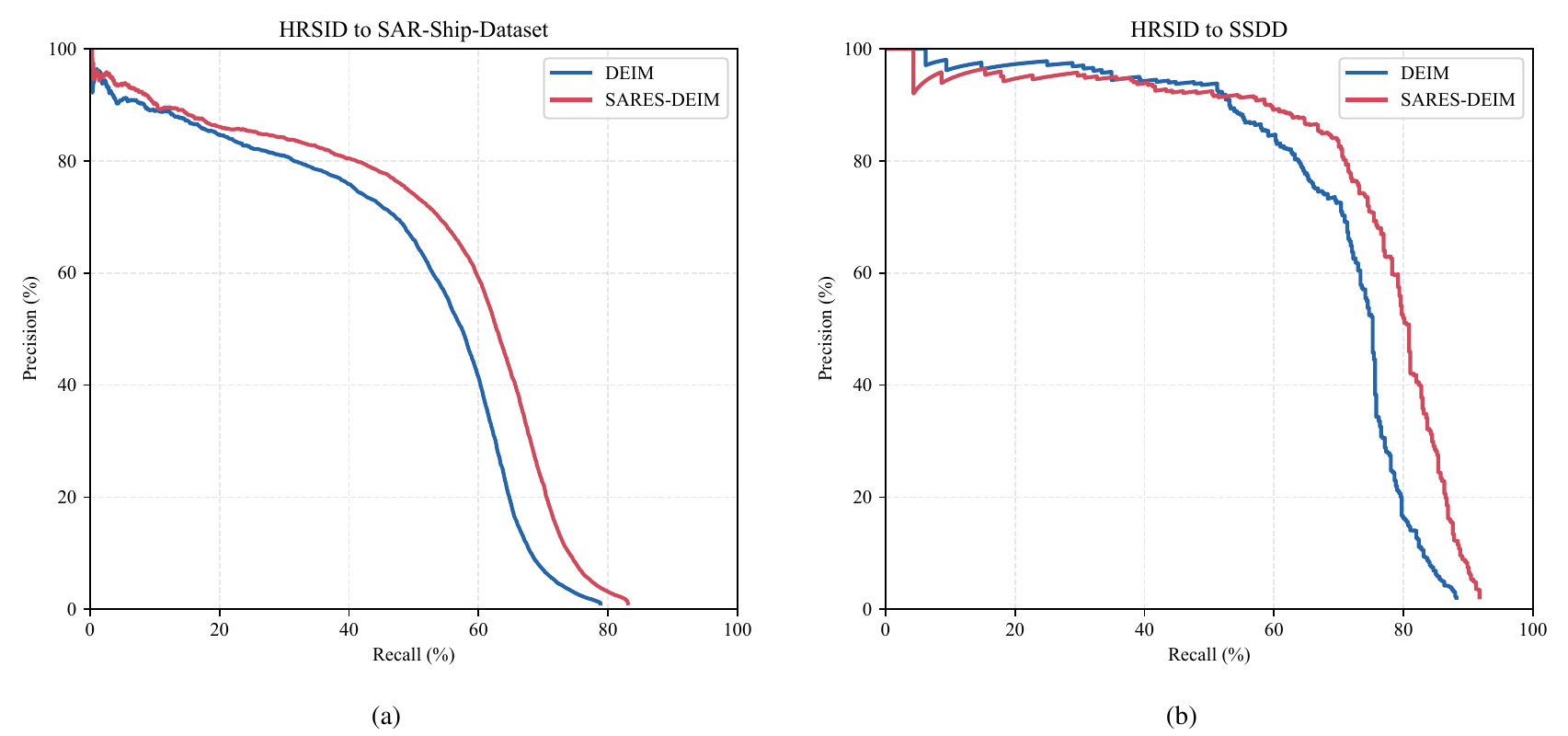}
  \caption{Confidence-ranked precision--recall curves after explicit category mapping: (a) HRSID~\cite{wei2020hrsid} $\rightarrow$ SAR-Ship-Dataset~\cite{wang2019sar} and (b) HRSID $\rightarrow$ SSDD~\cite{zhang2021ssdd}.}
  \label{fig:cross_dataset_pr}
\end{figure}

The complete confidence-ranked precision--recall curves in Fig.~\ref{fig:cross_dataset_pr} provide the threshold-independent view. On SAR-Ship-Dataset, the SARES-DEIM curve is above the DEIM-S~\cite{huang2025deim} curve across most of the evaluated recall range. On SSDD, SARES-DEIM provides a broader high-recall envelope, consistent with its higher $mAP_{50}$ and fixed-threshold Recall.

\begin{figure}[t]
  \centering
  \includegraphics[width=\columnwidth]{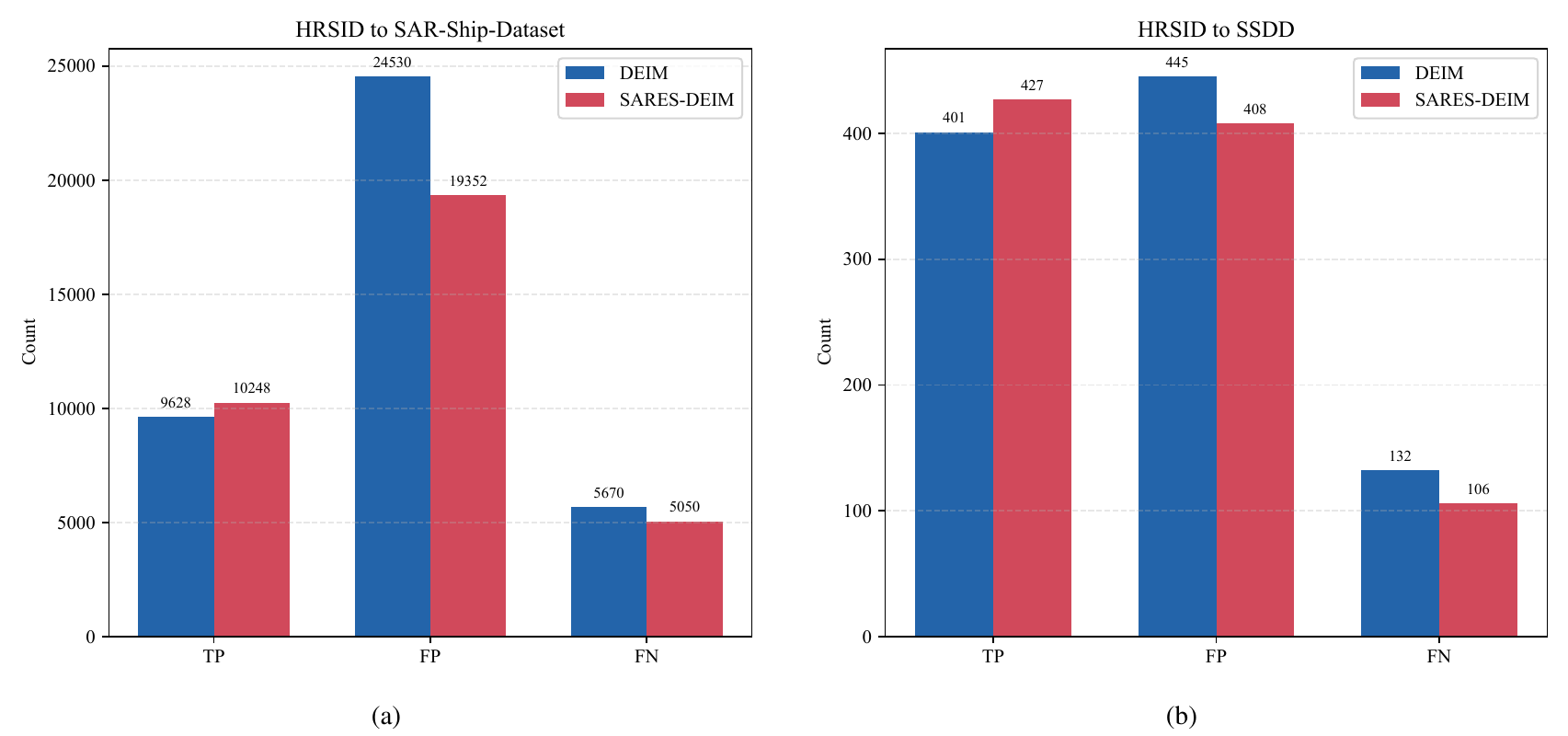}
  \caption{TP/FP/FN comparison at a confidence threshold of 0.25 and an IoU threshold of 0.50: (a) HRSID~\cite{wei2020hrsid} $\rightarrow$ SAR-Ship-Dataset~\cite{wang2019sar} and (b) HRSID $\rightarrow$ SSDD~\cite{zhang2021ssdd}.}
  \label{fig:cross_dataset_errors}
\end{figure}

The fixed-threshold error breakdown in Fig.~\ref{fig:cross_dataset_errors} provides a direct count-based check. SARES-DEIM does not obtain higher Precision by discarding true positives; it produces more true positives, fewer false positives, and fewer false negatives on both target datasets at confidence 0.25.

Taken together, the shared-threshold results, precision--recall curves, and error counts show simultaneous Precision and Recall improvements rather than a Precision--Recall trade-off after explicit category mapping. Nevertheless, the low absolute source-only AP, particularly on SAR-Ship-Dataset, shows that considerable domain shift remains. We therefore use this experiment only to disclose source-only behavior and its limitations; the main evidence for SARES-DEIM remains the supervised in-domain evaluations.

\subsubsection{Scene-Wise Analysis on HRSID}

To further analyze performance under different maritime backgrounds, we evaluate the official inshore and offshore scene groups provided by HRSID. As shown in Table~\ref{tab:scene_group}, SARES-DEIM improves over DEIM-S~\cite{huang2025deim} in both subsets. The gain is more evident in inshore scenes, where coastal clutter is stronger: $mAP_{50:95}$ increases from 74.7\% to 75.8\%, and Recall improves from 86.3\% to 87.7\%. In offshore scenes, both methods already achieve high $mAP_{50}$ and Recall, while SARES-DEIM still improves $mAP_{50:95}$ from 90.3\% to 91.0\%.

\begin{table}[t]
  \caption{Scene-wise performance comparison on the official HRSID~\cite{wei2020hrsid} inshore/offshore subsets. All detection metrics are reported in percent, and bold indicates the better result within each scene group; ties are both bold. Base denotes the direct DEIM-S~\cite{huang2025deim} architectural baseline.}
  \label{tab:scene_group}
  \centering
  \footnotesize
  \renewcommand{\arraystretch}{1.15}
  \setlength{\tabcolsep}{2.5pt}
  \begin{tabular*}{\columnwidth}{@{\extracolsep{\fill}}llcccc@{}}
    \toprule
    Scenario & Method & $mAP_{50:95}$\,(\%) & $mAP_{50}$\,(\%) & $P$\,(\%) & $R$\,(\%) \\
    \midrule
    Inshore  & DEIM-S~\cite{huang2025deim} (Base) & 74.7 & 92.3 & 90.3 & 86.3 \\
    Inshore  & \textbf{SARES-DEIM} & \textbf{75.8} & \textbf{93.2} & \textbf{90.9} & \textbf{87.7} \\
    Offshore & DEIM-S~\cite{huang2025deim} (Base) & 90.3 & \textbf{99.4} & 98.8 & \textbf{98.6} \\
    Offshore & \textbf{SARES-DEIM} & \textbf{91.0} & \textbf{99.4} & \textbf{99.1} & \textbf{98.6} \\
    \bottomrule
  \end{tabular*}
\end{table}

\subsection{Ablation Studies and Analysis}\label{sec:ablation}

To assess the individual contributions of the proposed modules, we conduct ablation studies on the HRSID official dataset using the official split ratio and the same training and evaluation protocol as the main experiments. Unless stated otherwise, all ablation tables below follow this HRSID protocol and change only the factor named in the corresponding table. The module-level ablation explicitly varies the presence and placement of SARESMoE and SDEP. The remaining controlled ablations keep the default settings fixed unless they are the variable under study: the default expert bank is 4/8/16 at P3/P4/P5, the default sparse activation is Top-2, and SDEP is enabled except in experiments that explicitly remove or vary it. The analysis covers module-level contributions, routing design, expert composition, expert-bank size, Top-$k$ sparsity, and SDEP insertion level.

\subsubsection{Module-Level Ablation}\label{sec:module_ablation}

We then evaluate the individual and combined contributions of SARESMoE and SDEP at the module level through incremental integration experiments, as summarized in Table~\ref{tab:ablation}.

\noindent\textbf{Effectiveness of the SDEP Neck.}
Integrating SDEP into the baseline without any MoE modules (SDEP~$\checkmark$, all MoE~$\times$, vs.\ the pure baseline) improves $mAP_{50:95}$ from 73.0\% to 74.9\% and Recall from 84.4\% to 87.3\%. The scale-wise metrics further show that $AP_S$ increases from 73.5\% to 76.2\%, suggesting that explicitly harvesting high-resolution cues from P2 helps preserve local structural information for small SAR ships.

\noindent\textbf{Impact of SARESMoE and its Hierarchical Placement.}
Introducing SARESMoE at P3 without SDEP improves $mAP_{50:95}$ from 73.0\% to 74.7\%. Extending SARESMoE to all P3/P4/P5 stages further improves the score to 75.3\%. The intermediate P3+P4 configuration performs lower, indicating that partial expert insertion does not necessarily yield monotonic gains and that the placement of expert modules matters.

\noindent\textbf{Synergy between SDEP and SARESMoE.}
The full \textsc{SARES-DEIM} model (SDEP~$\checkmark$, all MoE stages~$\checkmark$) achieves the highest $mAP_{50:95}$ of 76.4\% and the highest $AP_S$ of 77.0\% among the listed configurations. Compared to using SARESMoE alone or SDEP alone, the full model provides additional gains, suggesting that SDEP and SARESMoE play complementary roles in detail preservation and clutter-aware feature refinement.

\begin{table*}[t]
  \caption{Module-level ablation on the HRSID dataset~\cite{wei2020hrsid} with scale-specific AP metrics. The SARESMoE module is incrementally applied to the spatial pyramid stages (P3, P4, and P5). Bold indicates the best value in each detection-metric column.}
  \label{tab:ablation}
  \centering
  \footnotesize
  \renewcommand{\arraystretch}{1.15}
  \setlength{\tabcolsep}{2.2pt}
  \begin{tabular*}{\linewidth}{@{\extracolsep{\fill}}ccccccccccc@{}}
    \toprule
    \multirow{2}{*}{SDEP}
      & \multicolumn{3}{c}{SARESMoE Stages}
      & \multirow{2}{*}{$mAP_{50:95}$\,(\%)}
      & \multirow{2}{*}{$mAP_{50}$\,(\%)}
      & \multirow{2}{*}{$AP_S$\,(\%)}
      & \multirow{2}{*}{$AP_M$\,(\%)}
      & \multirow{2}{*}{$AP_L$\,(\%)}
      & \multirow{2}{*}{$P$\,(\%)}
      & \multirow{2}{*}{$R$\,(\%)} \\
    \cmidrule(lr){2-4}
      & P3 & P4 & P5 & & & & & & & \\
    \midrule
    $\times$ & $\times$ & $\times$ & $\times$ & 73.0 & 92.0 & 73.5 & 75.0 & 32.5 & 92.4 & 84.4 \\
    $\times$ & $\checkmark$ & $\times$ & $\times$ & 74.7 & 92.9 & 76.0 & 76.2 & 34.1 & 92.5 & 87.4 \\
    $\times$ & $\checkmark$ & $\checkmark$ & $\times$ & 72.9 & 92.8 & 73.6 & 72.7 & 31.0 & 91.4 & 85.9 \\
    $\times$ & $\checkmark$ & $\checkmark$ & $\checkmark$ & 75.3 & 93.6 & 76.0 & 76.2 & 34.3 & 92.0 & 87.1 \\
    \midrule
    $\checkmark$ & $\times$ & $\times$ & $\times$ & 74.9 & 92.9 & 76.2 & 76.3 & 33.1 & \textbf{93.1} & 87.3 \\
    $\checkmark$ & $\checkmark$ & $\times$ & $\times$ & 75.6 & \textbf{93.8} & 76.3 & 76.1 & \textbf{36.0} & 93.0 & 87.4 \\
    $\checkmark$ & $\checkmark$ & $\checkmark$ & $\times$ & 75.7 & 93.7 & 76.4 & 76.2 & 32.1 & 92.5 & 87.6 \\
    $\checkmark$ & $\checkmark$ & $\checkmark$ & $\checkmark$
      & \textbf{76.4} & \textbf{93.8} & \textbf{77.0} & \textbf{76.8} & 34.6 & \textbf{93.1} & \textbf{88.0} \\
    \bottomrule
  \end{tabular*}
\end{table*}

\subsubsection{SAR-Aware Routing Mechanism Analysis}\label{sec:router_ablation}

The core of SARESMoE is its routing mechanism, which determines sparse expert activation based on pooled image-level and pyramid-level spatial-frequency context. Standard MoE networks typically employ simple Multi-Layer Perceptrons (MLPs) for routing; however, SAR imagery may benefit from jointly considering spatial and frequency-domain cues. We therefore conduct an ablation study on the router architecture, as detailed in Table~\ref{tab:router_ablation}. In this experiment, the heterogeneous expert composition and 4/8/16 expert-bank setting are kept fixed, and only the learnable router architecture is changed; all variants use Top-2 sparse activation. Here, ``Spatial Branch'' and ``Frequency Branch'' denote the cue-extraction branches used to compute routing weights; they are distinct from the sparse expert types analyzed in Table~\ref{tab:expert_ablation}.

\noindent\textbf{Impact of Generic Routing.}
The \emph{Standard MLP Router} obtains 75.2\% $mAP_{50:95}$ and 86.6\% Precision, the lowest values among the learnable router variants. This suggests that generic MLP routing may not provide enough SAR-oriented inductive bias for separating targets from clutter.

\noindent\textbf{Single-Branch vs.\ Dual-Branch Analysis.}
We further dismantle our proposed SAR-aware router into its constituent cue branches. Using the \emph{Frequency Branch Only} or the \emph{Spatial Branch Only} yields $mAP_{50:95}$ scores of 75.3\% and 75.7\%, respectively. While these single-domain routers outperform the MLP baseline, their Precision remains lower than that of the dual-branch design. In contrast, the proposed \emph{Dual-Branch Fusion} integrates both frequency and spatial analyses, achieving the highest overall $mAP_{50:95}$ (76.4\%) and $AP_S$ (77.0\%) among the tested routing strategies. These results suggest that combining spectral and spatial cues is beneficial for expert assignment in heterogeneous maritime scenes.

\noindent\textbf{Routing Policy Comparison.}
To further isolate learned routing from the manually designed heterogeneous expert-bank allocation, we compare three distinct policies under the same 4/8/16 configuration. Uniform All-Expert Aggregation bypasses the router, assigns every stored sparse expert an equal weight $1/N_e$, and executes the full expert bank. Fixed Top-2 Routing also bypasses the router, but executes only the first two statically assigned experts with equal weights. Dynamic Top-2 Routing retains the learned dual-branch router, which selects and renormalizes two experts for each input and pyramid level. The results are summarized in the routing-policy table.

As shown in Table~\ref{tab:routing_policy}, Dynamic Top-2 Routing is 1.0 percentage point higher than Fixed Top-2 Routing in $mAP_{50:95}$ (76.4\% vs.\ 75.4\%) in the seed-0 runs, while maintaining a sparse active-expert regime. Uniform All-Expert Aggregation activates the entire expert bank and reaches 128.8 GFLOPs, but obtains lower $mAP_{50:95}$ than Dynamic Top-2 Routing (75.6\% vs.\ 76.4\%). The difference is consistent with their implementation: Uniform All-Expert Aggregation uses equal weights, whereas Top-All in Table~\ref{tab:topk_sparsity} keeps the learned router active and uses its input-dependent weights while executing all experts. We therefore interpret the seed-0 comparison as evidence for the value of adaptive weighting and sparse selection, rather than as a capacity-only effect.

\noindent\textbf{Multi-Seed Robustness.}
We repeated Fixed Top-2 Routing and Dynamic Top-2 Routing using seeds 13, 21, and 42 under the same protocol as seed 0. The multi-seed table reports every run and the sample standard deviation across the four seeds. Dynamic Top-2 Routing exceeds Fixed Top-2 Routing in $mAP_{50:95}$ for every matched seed. The paired gains for seeds 0, 13, 21, and 42 are 1.0, 1.2, 1.3, and 0.9 percentage points, respectively, giving a mean paired gain of $1.10\pm0.18$ percentage points. This consistency indicates that the observed improvement is not specific to seed 0.

\begin{table*}[t]
  \caption{Ablation study of distinct learnable router architectures within SARESMoE on HRSID~\cite{wei2020hrsid} with scale-specific AP metrics. Bold indicates the best performance in each detection-metric column.}
  \label{tab:router_ablation}
  \centering
  \footnotesize
  \renewcommand{\arraystretch}{1.15}
  \setlength{\tabcolsep}{2.7pt}
  \begin{tabular*}{\linewidth}{@{\extracolsep{\fill}}lccccccc@{}}
    \toprule
    Routing Strategy & $mAP_{50:95}$\,(\%) & $mAP_{50}$\,(\%) & $AP_S$\,(\%) & $AP_M$\,(\%) & $AP_L$\,(\%) & $P$\,(\%) & $R$\,(\%) \\
    \midrule
    Standard MLP Router                & 75.2 & 93.4 & 75.8 & 76.2 & 24.6 & 86.6 & 89.6 \\
    Frequency Branch Only              & 75.3 & 93.6 & 76.2 & 75.2 & 20.6 & 87.7 & 89.9 \\
    Spatial Branch Only                & 75.7 & \textbf{93.8} & 76.4 & 75.8 & 28.7 & 87.9 & \textbf{90.2} \\
    \textbf{Dual-Branch Fusion (Ours)} & \textbf{76.4} & \textbf{93.8} & \textbf{77.0} & \textbf{76.8} & \textbf{34.6} & \textbf{93.1} & 88.0 \\
    \bottomrule
  \end{tabular*}
\end{table*}

\begin{table*}[t]
  \caption{Comparison of routing policies under the 4/8/16 heterogeneous expert-bank configuration on HRSID~\cite{wei2020hrsid}. The routing policies are defined in the accompanying text; the table reports only quantitative results. Bold indicates the best value in each reported column; lower is better for Params and FLOPs.}
  \label{tab:routing_policy}
  \centering
  \footnotesize
  \renewcommand{\arraystretch}{1.15}
  \setlength{\tabcolsep}{2.7pt}
  \begin{tabular*}{\linewidth}{@{\extracolsep{\fill}}lcccccc@{}}
    \toprule
    Variant & $mAP_{50:95}$\,(\%) & $mAP_{50}$\,(\%) & $P$\,(\%) & $R$\,(\%) & Params (M) & FLOPs (G) \\
    \midrule
    Uniform All-Expert Aggregation & 75.6 & 93.4 & \textbf{93.1} & 87.9 & \textbf{87.6} & 128.8 \\
    Fixed Top-2 Routing & 75.4 & 93.2 & 92.1 & \textbf{88.4} & 90.7 & 75.3 \\
    \textbf{Dynamic Top-2 Routing (Ours)} & \textbf{76.4} & \textbf{93.8} & \textbf{93.1} & 88.0 & 90.8 & \textbf{72.6} \\
    \bottomrule
  \end{tabular*}
\end{table*}

\begin{table*}[t]
  \caption{Four-seed comparison of Fixed Top-2 Routing and Dynamic Top-2 Routing on HRSID~\cite{wei2020hrsid}. Mean $\pm$ sample standard deviation is reported in percentage points; bold indicates the better mean in each detection-metric column.}
  \label{tab:routing_multiseed}
  \centering
  \footnotesize
  \renewcommand{\arraystretch}{1.10}
  \setlength{\tabcolsep}{4pt}
  \begin{tabular*}{\linewidth}{@{\extracolsep{\fill}}llcccc@{}}
    \toprule
    Routing Policy & Seed & $mAP_{50:95}$\,(\%) & $mAP_{50}$\,(\%) & $P$\,(\%) & $R$\,(\%) \\
    \midrule
    \multirow{5}{*}{Fixed Top-2 Routing}
      & 0 & 75.4 & 93.2 & 92.1 & 88.4 \\
      & 13 & 75.1 & 92.8 & 93.2 & 87.3 \\
      & 21 & 75.1 & 93.4 & 92.5 & 88.4 \\
      & 42 & 75.4 & 93.0 & 93.1 & 87.7 \\
      & Mean $\pm$ SD & $75.25\pm0.17$ & $93.10\pm0.26$ & $92.73\pm0.52$ & \textbf{$87.95\pm0.54$} \\
    \midrule
    \multirow{5}{*}{Dynamic Top-2 Routing}
      & 0 & 76.4 & 93.8 & 93.1 & 88.0 \\
      & 13 & 76.3 & 94.0 & 92.8 & 87.8 \\
      & 21 & 76.4 & 93.9 & 93.1 & 86.7 \\
      & 42 & 76.3 & 93.7 & 92.0 & 87.5 \\
      & Mean $\pm$ SD & $\mathbf{76.35\pm0.06}$ & $\mathbf{93.85\pm0.13}$ & \textbf{$92.75\pm0.52$} & $87.50\pm0.57$ \\
    \bottomrule
  \end{tabular*}
\end{table*}

\subsubsection{SDEP Fusion Strategy Analysis}\label{sec:sdep_ablation}

To further examine the architectural design of the SDEP neck, we investigate different downsampling and fusion strategies when injecting high-resolution features from the P2 layer into the P3 semantic level. All experiments in this subsection are built upon the default SARESMoE baseline (P3/P4/P5 with the 4/8/16 expert bank and Top-2 sparse activation), so only the P2-to-P3 fusion mechanism is varied. The results are summarized in Table~\ref{tab:sdep_fusion}.

\noindent\textbf{Impact of Lossy Downsampling.}
When naive fusion mechanisms are employed, such as \emph{Standard Strided Downsampling} or \emph{Strip Convolution}~\cite{yuan2025strip}, Recall increases slightly but Precision decreases from 92.0\% to 86.3\% and 86.8\%, respectively. The overall $mAP_{50:95}$ also drops below the SARESMoE baseline (75.3\%). This suggests that directly compressing P2 features with lossy or mismatched transformations can introduce feature aliasing and increase false positives.

\noindent\textbf{Effect of Space-to-Depth (SDEP).}
In contrast, SDEP employs a Space-to-Depth (SPD) transformation to rearrange shallow spatial pixels into the channel dimension. As shown in Table~\ref{tab:sdep_fusion}, the SDEP configuration achieves the highest $mAP_{50:95}$ (76.4\%) and improves $AP_S$ from 76.0\% to 77.0\% over the no-fusion baseline. This supports the use of space-to-depth fusion for preserving fine-grained spatial cues from shallow features.

\begin{table*}[t]
  \caption{Comparison of different P2-to-P3 fusion strategies on HRSID~\cite{wei2020hrsid} with scale-specific AP metrics. All configurations are evaluated on top of the SARESMoE baseline. Bold indicates the best performance in each detection-metric column.}
  \label{tab:sdep_fusion}
  \centering
  \footnotesize
  \renewcommand{\arraystretch}{1.15}
  \setlength{\tabcolsep}{2.7pt}
  \begin{tabular*}{\linewidth}{@{\extracolsep{\fill}}lccccccc@{}}
    \toprule
    Fusion Strategy (P2 $\rightarrow$ P3) & $mAP_{50:95}$\,(\%) & $mAP_{50}$\,(\%) & $AP_S$\,(\%) & $AP_M$\,(\%) & $AP_L$\,(\%) & $P$\,(\%) & $R$\,(\%) \\
    \midrule
    Baseline (No P2 Fusion)          & 75.3 & 93.6 & 76.0 & 76.2 & 34.3 & 92.0 & 87.1 \\
    Standard Strided Downsampling    & 74.4 & 93.2 & 75.3 & 74.4 & 24.3 & 86.3 & \textbf{89.6} \\
    Strip Convolution~\cite{yuan2025strip} & 74.1 & 93.2 & 75.0 & 74.7 & 28.1 & 86.8 & 89.4 \\
    \textbf{SDEP (Space-to-Depth)}   & \textbf{76.4} & \textbf{93.8} & \textbf{77.0} & \textbf{76.8} & \textbf{34.6} & \textbf{93.1} & 88.0 \\
    \bottomrule
  \end{tabular*}
\end{table*}

\subsubsection{Expert Composition Analysis}\label{sec:expert_ablation}

We first investigate how different expert compositions within SARESMoE influence detection performance. As shown in Table~\ref{tab:expert_ablation}, we conduct a series of controlled experiments in which individual expert types are isolated or replaced at specific pyramid levels. All configurations are evaluated without SDEP to isolate the effect of expert composition. Across all configurations, the Shared Expert remains always activated and is not governed by the routing mechanism.

\noindent\textbf{Homogeneous Shared Expert Configuration.}
When all sparse experts are globally replaced with the Shared Expert, the model achieves 74.5\% $mAP_{50:95}$. This homogeneous configuration remains below the full SARESMoE configuration (75.3\%), suggesting that heterogeneous experts provide useful complementary representations across semantic levels.

\noindent\textbf{Expert Specialization at P3.}
The P3 level processes the shallowest, highest-resolution features used by SARESMoE, where small ships are vulnerable to speckle corruption and local structural cues remain important. The Wavelet and Spatial Experts therefore play complementary roles: the former emphasizes wavelet-domain denoising, while the latter preserves compact target structures. Configuring P3 with only the Spatial Expert leads to a 1.5 percentage-point drop in $mAP_{50:95}$ compared with the full SARESMoE configuration. Employing only the Wavelet Expert at P3 recovers part of the performance and yields higher Recall, suggesting that wavelet-domain processing is useful for shallow-stage denoising. The full P3 configuration allows the router to allocate features between both expert types and achieves a better overall balance.

\noindent\textbf{Expert Specialization at P4 and P5.}
At the deeper P4 and P5 stages, we evaluate the individual contributions of the Frequency and Hybrid experts. At P4, the Frequency-only variant yields 73.9\% $mAP_{50:95}$, whereas the Hybrid-only variant achieves 74.2\%. At P5, the Frequency-only variant achieves 74.6\%, slightly above the Hybrid-only variant (73.8\%). These results indicate that expert preferences vary across pyramid levels, motivating the heterogeneous expert design.

\begin{table*}[t]
  \caption{Expert composition ablation on HRSID~\cite{wei2020hrsid} with scale-specific AP metrics. All models are evaluated without SDEP to isolate the MoE mechanism. The Shared Expert is uniformly active across all runs. For single-expert ablation rows, the specified level utilizes only the designated expert, while the unmentioned pyramid levels retain the default full SARESMoE expert configuration. Bold indicates the best performance in each detection-metric column.}
  \label{tab:expert_ablation}
  \centering
  \footnotesize
  \renewcommand{\arraystretch}{1.15}
  \setlength{\tabcolsep}{2.7pt}
  \begin{tabular*}{\linewidth}{@{\extracolsep{\fill}}lccccccc@{}}
    \toprule
    Configuration & $mAP_{50:95}$\,(\%) & $mAP_{50}$\,(\%) & $AP_S$\,(\%) & $AP_M$\,(\%) & $AP_L$\,(\%) & $P$\,(\%) & $R$\,(\%) \\
    \midrule
    Shared Expert Only        & 74.5 & 92.7 & 75.8 & 76.7 & 32.2 & 92.1 & 86.6 \\
    \midrule
    P3: Spatial Expert Only                & 73.8 & 92.4 & 75.2 & 75.3 & 28.6 & 92.8 & 85.6 \\
    P3: Wavelet Expert Only                & 74.1 & 92.7 & 75.4 & 75.8 & 33.3 & 92.5 & \textbf{87.3} \\
    \midrule
    P4: Frequency Expert Only              & 73.9 & 92.3 & 75.1 & 75.2 & \textbf{36.9} & 91.9 & 86.9 \\
    P4: Hybrid Expert Only                 & 74.2 & 92.8 & 75.6 & 74.9 & 29.6 & 92.3 & 86.6 \\
    \midrule
    P5: Frequency Expert Only              & 74.6 & 92.6 & 75.7 & \textbf{77.1} & 31.1 & \textbf{92.9} & 86.8 \\
    P5: Hybrid Expert Only                 & 73.8 & 92.4 & 75.2 & 75.0 & 33.4 & 92.2 & 87.1 \\
    \midrule
    \textbf{Full SARESMoE Module (w/o SDEP)} & \textbf{75.3} & \textbf{93.6} & \textbf{76.0} & 76.2 & 34.3 & 92.0 & 87.1 \\
    \bottomrule
  \end{tabular*}
\end{table*}

\subsubsection{Expert Number and Sparse Activation Analysis}\label{sec:expert_number_topk}

We further analyze two MoE-specific hyperparameters: the number of stored experts and the number of activated experts. Table~\ref{tab:expert_number} varies the expert-bank size while keeping SDEP enabled, the heterogeneous expert composition unchanged, and Top-2 sparse activation. Merely increasing the number of stored experts does not consistently improve detection accuracy. The progressive 4/8/16 configuration achieves the highest $mAP_{50:95}$ among the tested settings. Because Top-$k$ is fixed in this table, the reported FLOPs are computed from the same number of active sparse-expert paths per input rather than from all stored experts; accordingly, they remain comparable across expert-bank sizes, while Params mainly reflect the storage capacity of the expert bank.

\begin{table*}[t]
  \caption{Ablation study of SARESMoE expert number on HRSID~\cite{wei2020hrsid}. All settings use Top-2 sparse activation; only the number of stored experts changes. Bold indicates the best detection metric in each column.}
  \label{tab:expert_number}
  \centering
  \footnotesize
  \renewcommand{\arraystretch}{1.15}
  \setlength{\tabcolsep}{3.3pt}
  \begin{tabular*}{\linewidth}{@{\extracolsep{\fill}}lccccccc@{}}
    \toprule
    Expert Number & $mAP_{50:95}$\,(\%) & $mAP_{50}$\,(\%) & $AP_S$\,(\%) & $AP_M$\,(\%) & $AP_L$\,(\%) & Params (M) & FLOPs (G) \\
    \midrule
    2/2/2                & 74.7 & 93.1 & 75.9 & 75.9 & 33.5 & 37.6 & 72.6 \\
    4/4/4                & 74.8 & 92.9 & 75.9 & 76.5 & 32.1 & 46.4 & 72.6 \\
    8/8/8                & 75.1 & 93.0 & 76.4 & 76.0 & \textbf{35.5} & 63.9 & 72.6 \\
    16/16/16             & 74.8 & 93.0 & 76.1 & 75.6 & 30.2 & 99.1 & 72.6 \\
    \textbf{Ours (4/8/16)} & \textbf{76.4} & \textbf{93.8} & \textbf{77.0} & \textbf{76.8} & 34.6 & 90.8 & 72.6 \\
    \bottomrule
  \end{tabular*}
\end{table*}

Table~\ref{tab:topk_sparsity} evaluates Top-$k$ gating sparsity while keeping SDEP enabled and the default 4/8/16 expert bank unchanged. Top-1 has the lowest FLOPs but lower $mAP_{50:95}$ than the default setting. Top-All sets $k=N_e$ in the learned-router mode, so every expert is executed but the router's input-dependent Softmax weights are retained; it is therefore different from Uniform All-Expert Aggregation, which removes the router and averages all experts equally. Activating all experts increases FLOPs to 126.0 GFLOPs without improving accuracy. Top-2 provides the most favorable accuracy-computation trade-off in this experiment, supporting the use of sparse activation: the model can retain a larger expert bank for SAR-specific representation capacity while keeping the active per-sample computation controlled.

\begin{table*}[t]
  \caption{Ablation study of Top-$k$ gating sparsity in SARESMoE on HRSID~\cite{wei2020hrsid}. The expert bank is kept unchanged, while only the number of activated sparse experts is varied. Bold indicates the best detection metric in each column; lower FLOPs is preferred.}
  \label{tab:topk_sparsity}
  \centering
  \footnotesize
  \renewcommand{\arraystretch}{1.15}
  \setlength{\tabcolsep}{3.6pt}
  \begin{tabular*}{\linewidth}{@{\extracolsep{\fill}}lccccccc@{}}
    \toprule
    Gating Setting & $mAP_{50:95}$\,(\%) & $mAP_{50}$\,(\%) & $AP_S$\,(\%) & $AP_M$\,(\%) & $AP_L$\,(\%) & Params (M) & FLOPs (G) \\
    \midrule
    Top-1                 & 75.2 & 93.6 & 76.5 & 75.9 & \textbf{37.7} & 90.8 & \textbf{67.1} \\
    Top-All               & 75.0 & 92.9 & 76.2 & \textbf{76.8} & 35.4 & 90.8 & 126.0 \\
    \textbf{Top-2 (Ours)} & \textbf{76.4} & \textbf{93.8} & \textbf{77.0} & \textbf{76.8} & 34.6 & 90.8 & 72.6 \\
    \bottomrule
  \end{tabular*}
\end{table*}

\subsubsection{SDEP Insertion Level Analysis}\label{sec:sdep_level}

Finally, Table~\ref{tab:sdep_level} compares injecting high-resolution P2 features into different pyramid stages while keeping the default SARESMoE configuration and Top-2 sparse activation fixed. Injecting P2 features into P3 achieves the highest overall $mAP_{50:95}$ among the tested insertion levels and improves small-target AP compared with the no-SDEP setting. Direct injection into P4 or P5 yields lower overall AP, suggesting that P3 is a suitable destination for preserving fine-grained spatial cues from P2 while limiting the resolution-semantic mismatch.

\begin{table*}[t]
  \caption{Ablation study of SDEP insertion levels on HRSID~\cite{wei2020hrsid}. P2 features are injected into different pyramid stages and evaluated with scale-specific AP metrics. Bold indicates the best value in each column; lower is better for Params and FLOPs.}
  \label{tab:sdep_level}
  \centering
  \footnotesize
  \renewcommand{\arraystretch}{1.15}
  \setlength{\tabcolsep}{3.8pt}
  \begin{tabular*}{\linewidth}{@{\extracolsep{\fill}}lccccccc@{}}
    \toprule
    SDEP Insertion Level & $mAP_{50:95}$\,(\%) & $mAP_{50}$\,(\%) & $AP_S$\,(\%) & $AP_M$\,(\%) & $AP_L$\,(\%) & Params (M) & FLOPs (G) \\
    \midrule
    No SDEP                          & 75.3 & 93.6 & 76.0 & 76.2 & 34.3 & \textbf{90.2} & \textbf{63.4} \\
    P2 $\rightarrow$ P4              & 74.1 & 92.8 & 75.3 & 75.9 & \textbf{41.8} & 92.9 & 68.4 \\
    P2 $\rightarrow$ P5              & 74.0 & 92.4 & 75.3 & 75.4 & 28.5 & 91.2 & 67.5 \\
    \textbf{P2 $\rightarrow$ P3 (Ours)} & \textbf{76.4} & \textbf{93.8} & \textbf{77.0} & \textbf{76.8} & 34.6 & 90.8 & 72.6 \\
    \bottomrule
  \end{tabular*}
\end{table*}

\subsection{Qualitative and Routing Analysis}\label{sec:vis}

We provide qualitative detection comparisons, expert-routing statistics, and CAM visualizations to complement the quantitative results. The analysis covers bounding-box comparisons (Sec.~\ref{sec:vis_det}), activation statistics (Sec.~\ref{sec:vis_activation_stats}), expert-level CAMs (Sec.~\ref{sec:vis_expert}), and module-level CAMs (Sec.~\ref{sec:vis_module}). These analyses are intended as empirical observations of model behavior, not as strict proof of physical interpretability.

\subsubsection{Qualitative Detection Comparison}\label{sec:vis_det}

\begin{figure}[t]
  \centering
  \includegraphics[width=0.95\linewidth]{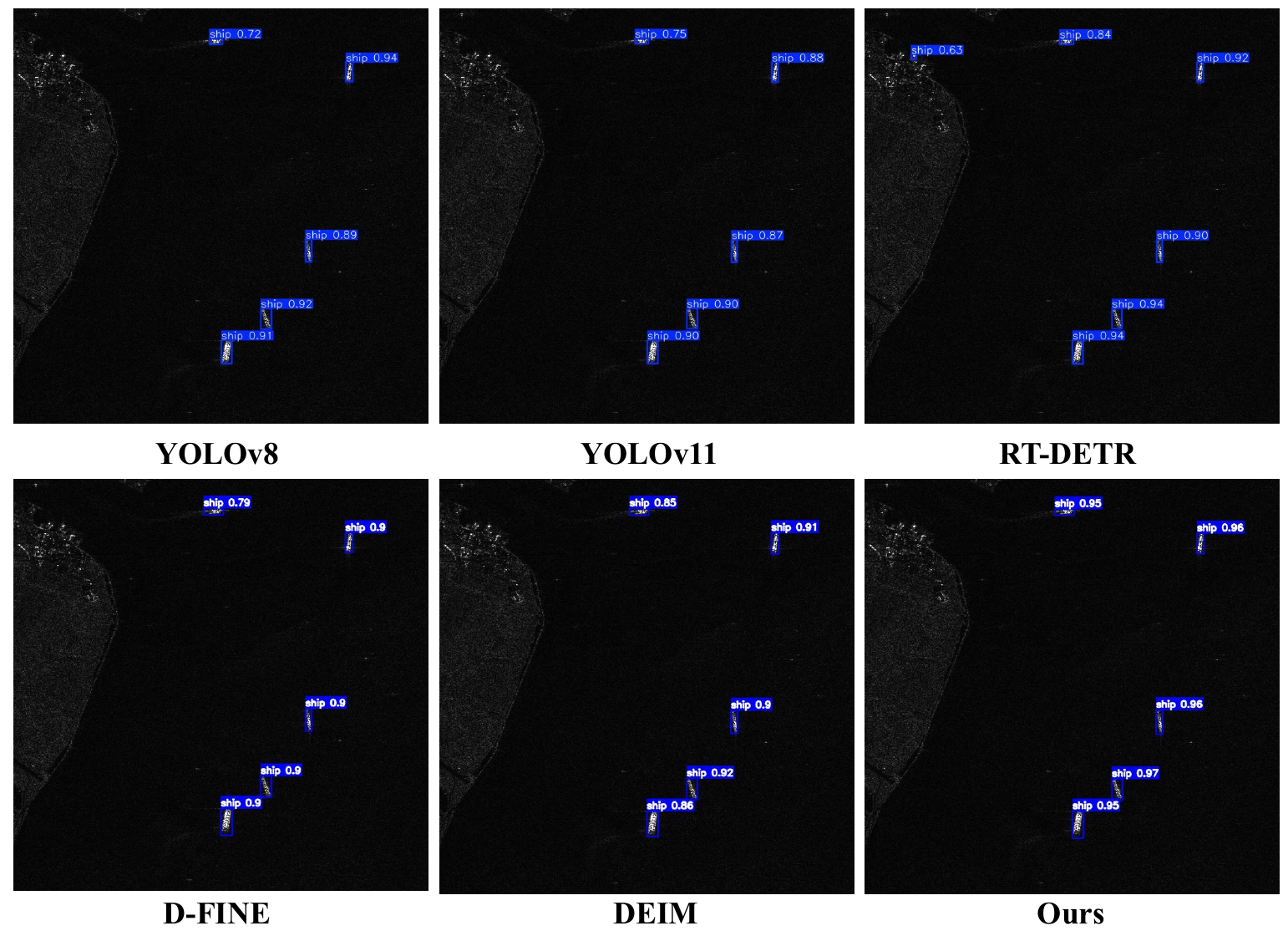}
  \caption{Qualitative bounding-box comparison on representative HRSID~\cite{wei2020hrsid} samples. The figure compares representative YOLO-series, DETR-series, and SARES-DEIM detection outputs under challenging SAR ship-detection scenarios.}
  \label{fig:hrsid_bbox_comparison}
\end{figure}

\begin{figure}[t]
  \centering
  \includegraphics[width=0.95\linewidth]{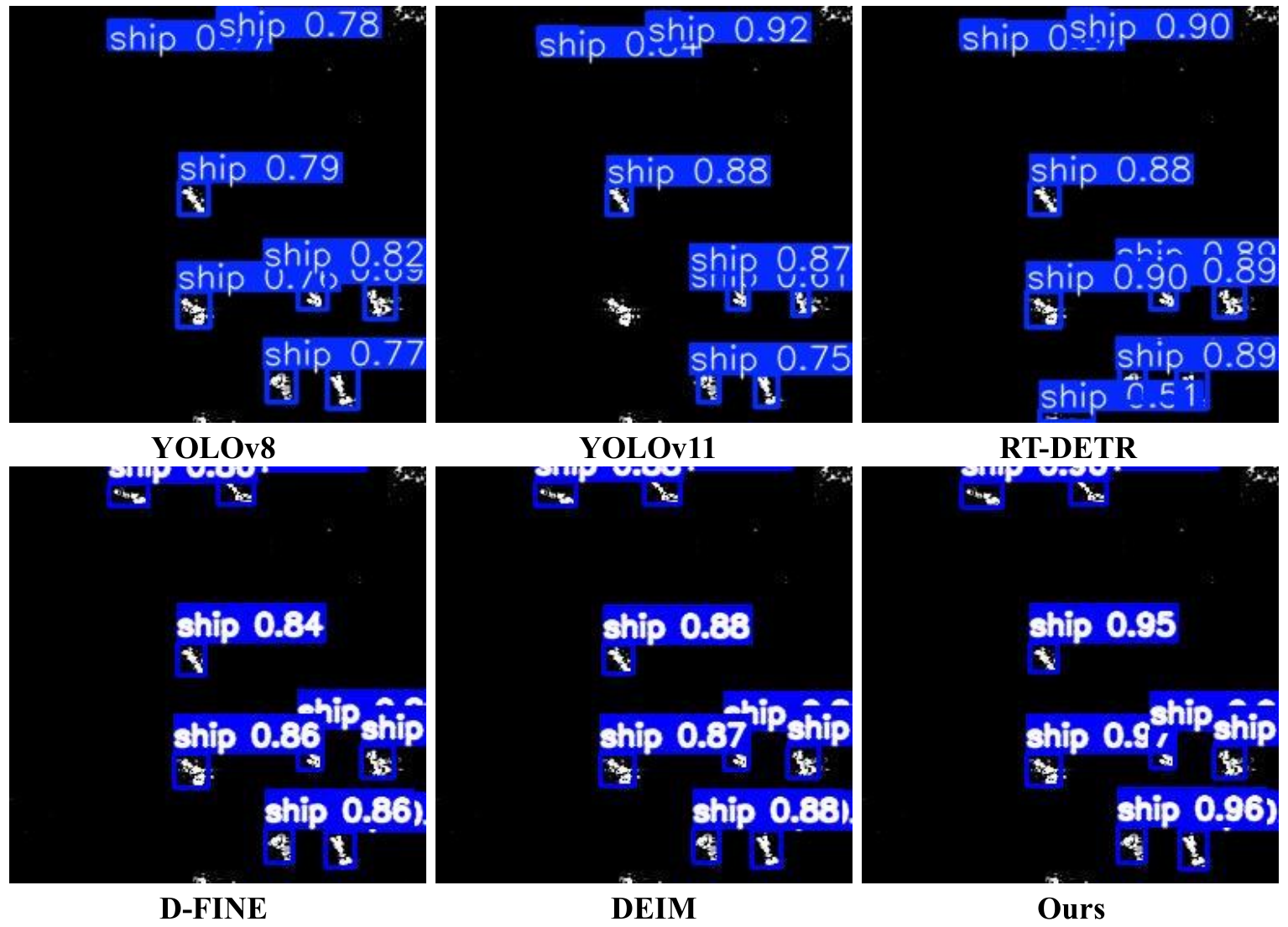}
  \caption{Qualitative bounding-box comparison on representative SAR-Ship-Dataset~\cite{wang2019sar} samples. The figure compares representative YOLO-series, DETR-series, and SARES-DEIM detection outputs under different SAR ship appearances and backgrounds.}
  \label{fig:sarship_bbox_comparison}
\end{figure}

Figs.~\ref{fig:hrsid_bbox_comparison} and~\ref{fig:sarship_bbox_comparison} compare bounding-box outputs of representative YOLO-series detectors, DETR-series detectors, and SARES-DEIM on SAR ship images. The HRSID samples include cluttered inshore scenes and small-scale targets, while the SAR-Ship-Dataset samples cover different ship appearances and sensor backgrounds. The qualitative results suggest that SARES-DEIM tends to produce tighter boxes around small ships and fewer clutter-induced false alarms in several near-shore examples, which is consistent with the scale-wise and scene-wise results reported above.

\begin{figure*}[t]
  \centering
  \includegraphics[width=0.95\linewidth]{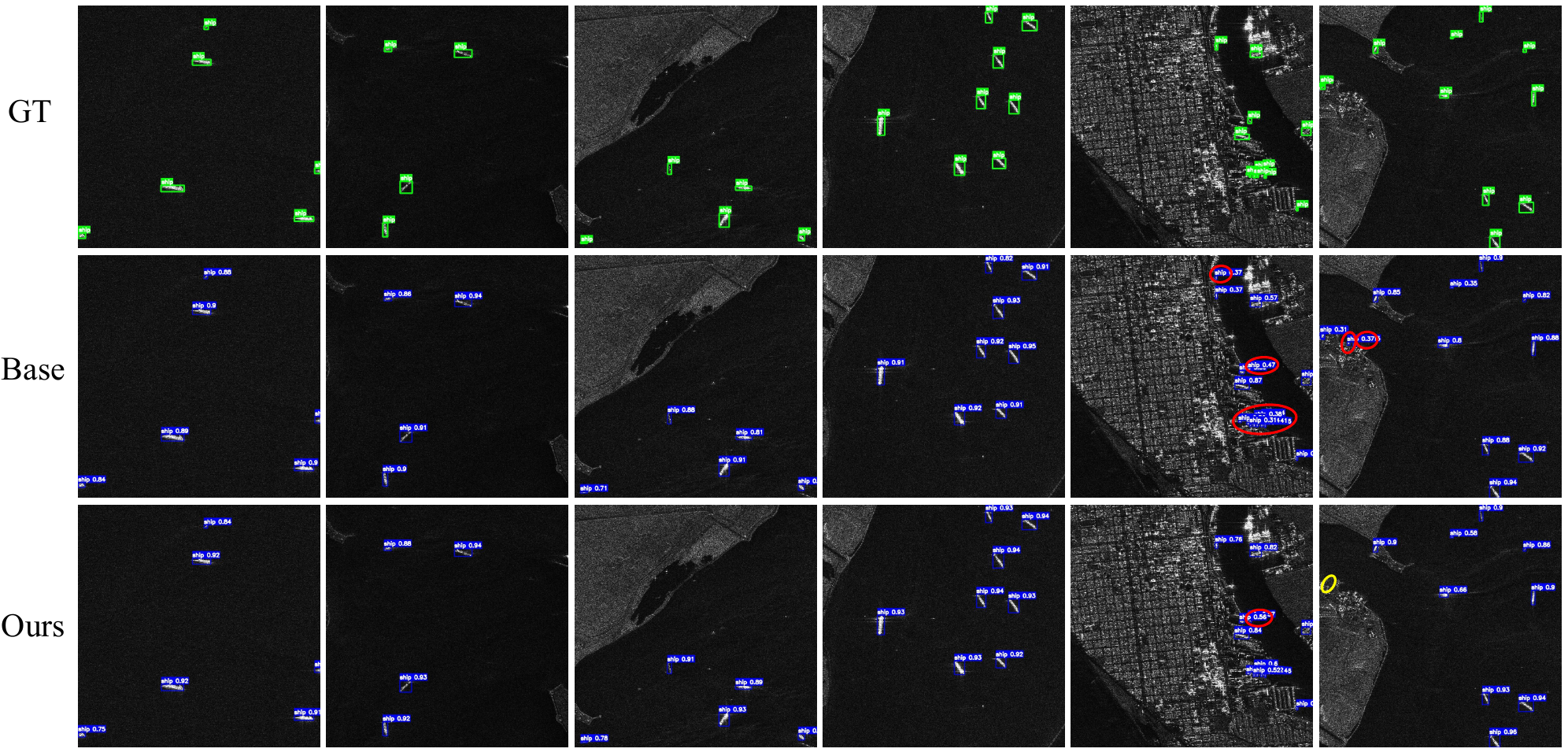}
  \caption{Qualitative comparison between the ground truth (GT), the DEIM-S baseline (Base), and the proposed SARES-DEIM (Ours) on representative HRSID~\cite{wei2020hrsid} samples. Yellow ellipses indicate missed detections, while red ellipses indicate false alarms. The focused comparison shows that SARES-DEIM can reduce several missed detections on small or weak targets and suppress some clutter-induced false positives compared with the baseline.}
  \label{fig:base_ours_bbox_comparison}
\end{figure*}

To further isolate the effect of the proposed modifications relative to the DEIM-S baseline, Fig.~\ref{fig:base_ours_bbox_comparison} provides a focused GT/Base/Ours comparison on representative HRSID samples. Yellow ellipses indicate missed detections, while red ellipses indicate false alarms. Compared with the baseline, SARES-DEIM reduces several missed detections on small or weak targets and suppresses some false alarms caused by coastal clutter or complex background structures, while challenging cases still remain.

\subsubsection{Expert Activation Statistics}\label{sec:vis_activation_stats}

To examine the behavior of the sparse router, we compute expert activation frequencies on the official inshore and offshore HRSID subsets used in Table~\ref{tab:scene_group}. Expert activation frequency is computed over Top-$k$ expert-selection events at each pyramid level and normalized within each level and scenario subset.

\begin{figure*}[t]
  \centering
  \includegraphics[width=0.92\linewidth]{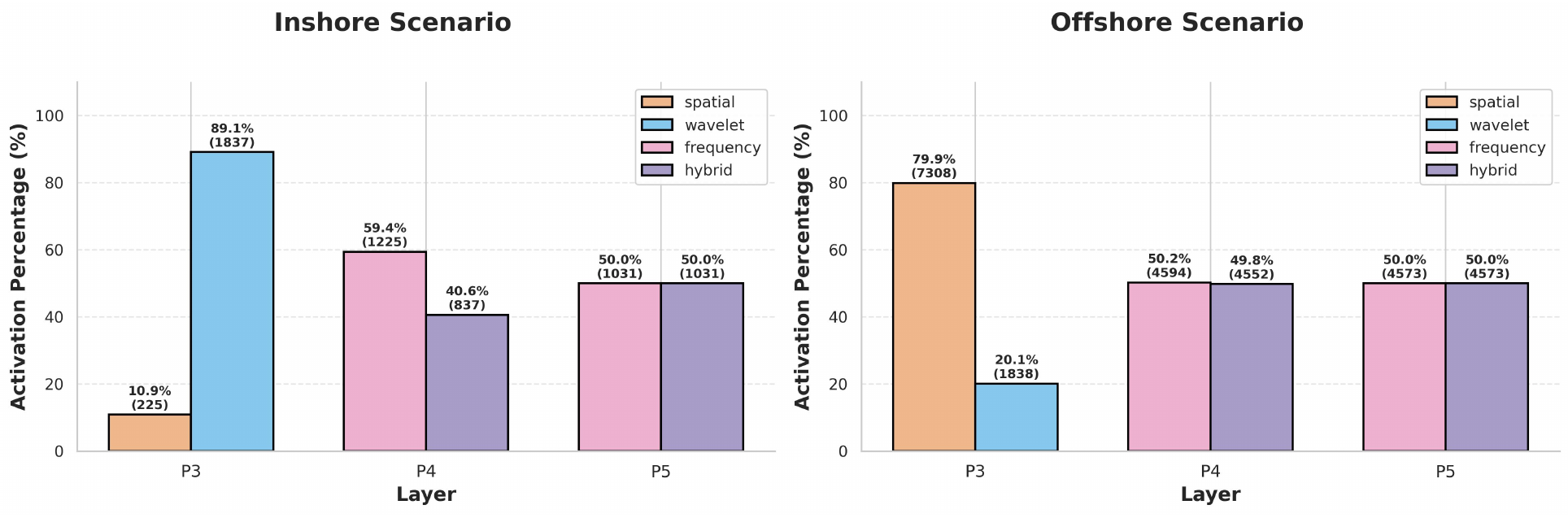}
  \caption{Expert-type activation statistics of SARESMoE on inshore and offshore HRSID~\cite{wei2020hrsid} subsets. The percentages indicate normalized frequencies of Top-$k$ sparse expert-selection events at P3, P4, and P5.}
  \label{fig:inshore_offshore_expert_activation}
\end{figure*}

As shown in Fig.~\ref{fig:inshore_offshore_expert_activation}, the shallow P3 router exhibits different preferences across maritime scenarios. In inshore scenes, P3 selects the Wavelet Expert much more frequently (89.1\%) than the Spatial Expert (10.9\%), suggesting that shallow features in cluttered coastal areas are often routed to the branch designed for denoising and high-frequency interference suppression. In offshore scenes, P3 instead favors the Spatial Expert (79.9\%), indicating that compact target-structure preservation is selected more frequently when the background is relatively homogeneous. At deeper P4 and P5 stages, Frequency and Hybrid experts are selected in a more balanced manner. We interpret these statistics as empirical routing behavior that supports the SAR-aware design, while avoiding a stronger claim of strict physical interpretability.

\begin{figure}[t]
  \centering
  \includegraphics[width=0.95\linewidth]{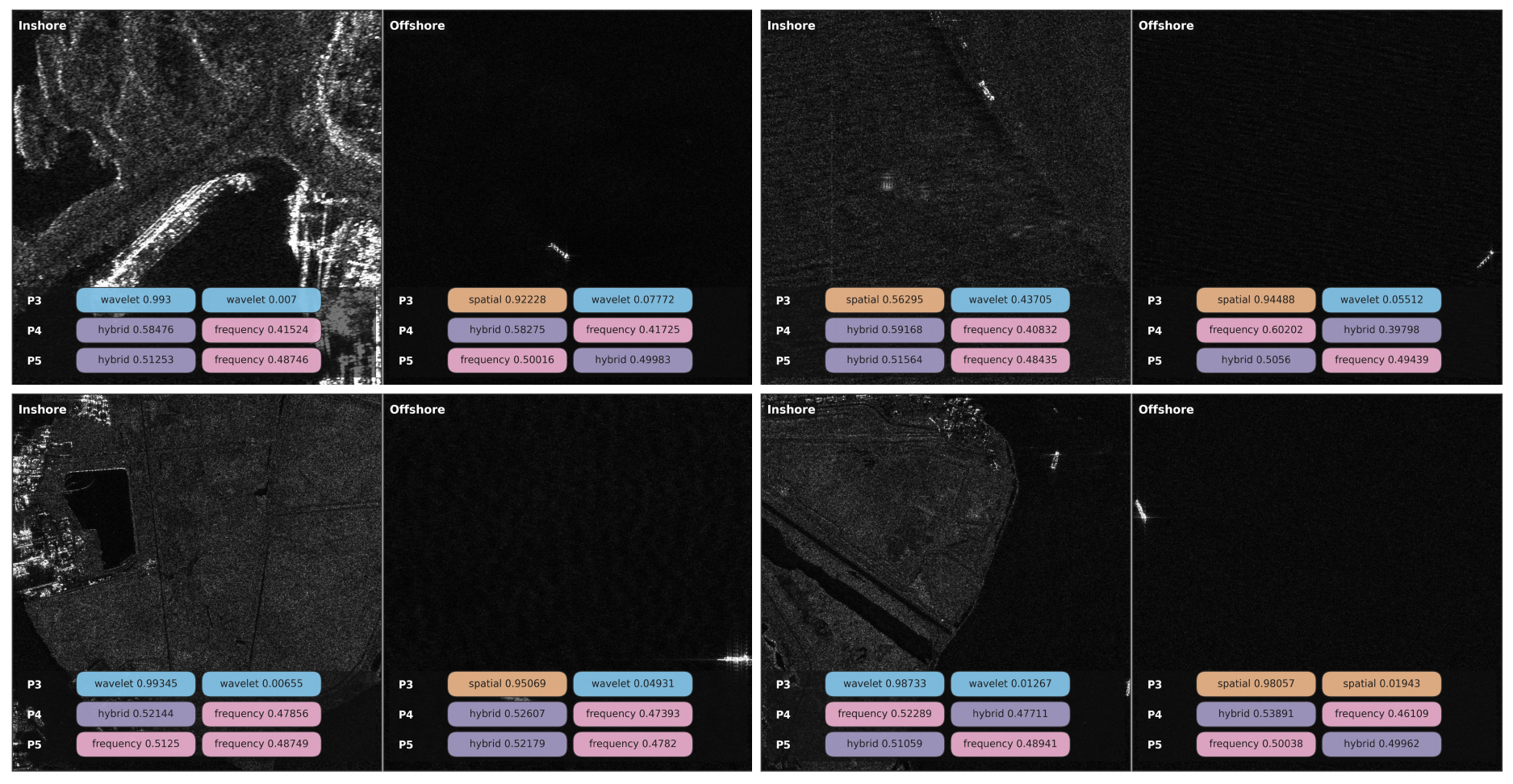}
  \caption{Representative paired routing cases on official HRSID~\cite{wei2020hrsid} inshore and offshore subsets. Each pair contains one inshore and one offshore SAR image, and the badges list the Top-$k$ selected sparse experts and their normalized routing weights at P3, P4, and P5. The visualization reflects image-level and pyramid-level expert selection rather than dense pixel-wise routing.}
  \label{fig:inshore_offshore_routing_cases}
\end{figure}

Fig.~\ref{fig:inshore_offshore_routing_cases} further provides representative paired cases from the official inshore/offshore subsets. These examples show the Top-$k$ expert selections and normalized routing weights at each pyramid level, complementing the aggregate statistics with image-level routing evidence.

\subsubsection{Expert-Level CAM Analysis}\label{sec:vis_expert}

\begin{figure}[t]
  \centering
  \includegraphics[width=0.92\linewidth]{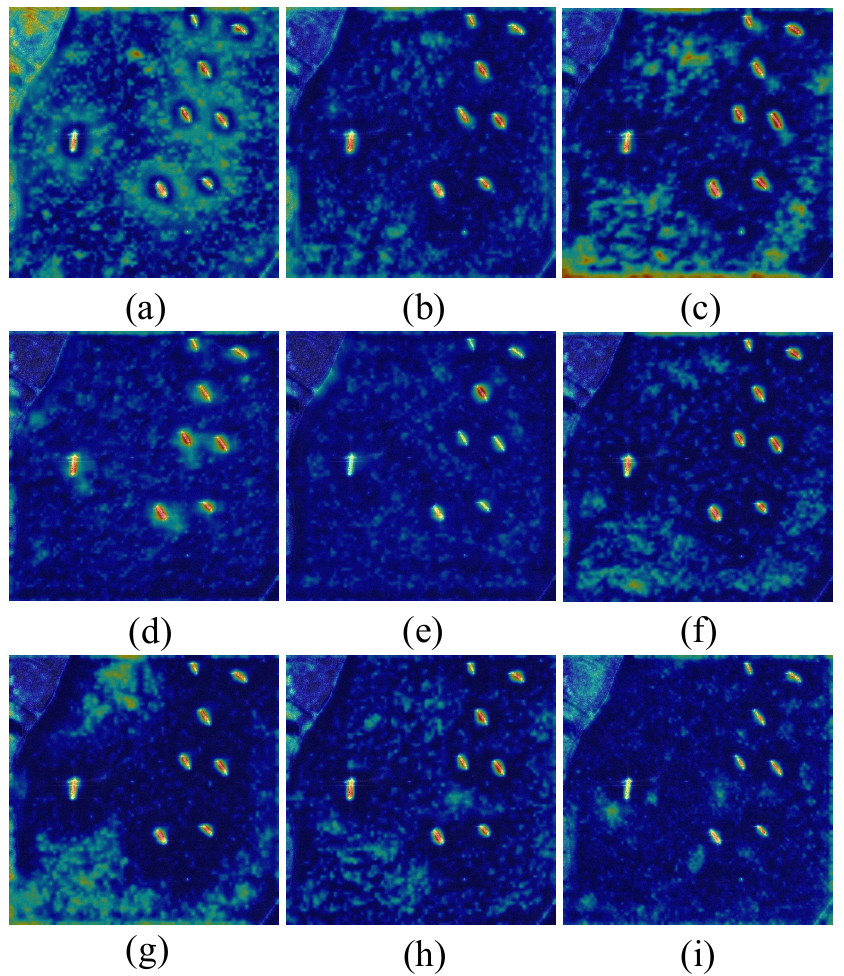}
  \caption{Expert-level CAM visualizations on HRSID~\cite{wei2020hrsid}. All configurations exclude SDEP: (a) all sparse experts replaced with the Shared Expert; (b) P3 Spatial only; (c) P3 Wavelet only; (d) P4 Frequency only; (e) P4 Hybrid only; (f) P5 Frequency only; (g) P5 Hybrid only; (h) Uniform All-Expert Aggregation; and (i) full SARESMoE. Unmentioned pyramid levels retain the default full SARESMoE expert configuration. Warmer colors indicate stronger feature responses.}
  \label{fig:vis_expert}
\end{figure}

To complement the routing-frequency statistics, we further inspect Class Activation Maps (CAMs) produced by representative expert configurations across pyramid levels in Fig.~\ref{fig:vis_expert}. This analysis is intended to provide qualitative evidence about where different expert variants tend to respond, rather than to establish strict physical interpretability. To isolate expert behavior from the detail-preserving contribution of SDEP, all nine configurations are visualized in a controlled SARESMoE-only setting without the SDEP neck. The panels cover the homogeneous Shared Expert control, individual Spatial/Wavelet/Frequency/Hybrid experts at P3/P4/P5, Uniform All-Expert Aggregation, and the full dynamically routed SARESMoE configuration.

\noindent\textbf{Shared Expert.}
When all sparse experts are replaced with the homogeneous Shared Expert, Fig.~\ref{fig:vis_expert}(a) shows a broad semantic response that covers the ship regions but also retains noticeable activation on cluttered near-shore structures. This behavior is consistent with the Shared Expert serving as a stable common feature extractor rather than a specialized clutter-suppression pathway.

\noindent\textbf{Expert Specialization at P3, P4, and P5.}
At P3, the Spatial-only configuration in Fig.~\ref{fig:vis_expert}(b) preserves local target boundaries, whereas the Wavelet-only configuration in Fig.~\ref{fig:vis_expert}(c) appears less diffuse in several displayed regions. At P4, the Frequency-only and Hybrid-only configurations in Fig.~\ref{fig:vis_expert}(d) and Fig.~\ref{fig:vis_expert}(e) exhibit complementary mid-level response patterns: the former emphasizes spectral filtering, while the latter retains additional structural responses through its multi-branch design. At P5, the Frequency-only response in Fig.~\ref{fig:vis_expert}(f) is more spatially concentrated in the displayed case, whereas the Hybrid-only response in Fig.~\ref{fig:vis_expert}(g) remains comparatively diffuse. These qualitative differences are consistent with the expert-composition ablation in Table~\ref{tab:expert_ablation}, which indicates that expert preferences vary across pyramid levels.

\noindent\textbf{Uniform All-Expert Aggregation vs. Full SARESMoE.}
Uniform All-Expert Aggregation in Fig.~\ref{fig:vis_expert}(h) removes input-dependent routing and assigns equal weights to every stored sparse expert. Its response retains more background activation and exhibits weaker target concentration in the displayed example. In contrast, the full SARESMoE configuration in Fig.~\ref{fig:vis_expert}(i) dynamically selects and weights the Top-2 sparse experts at each pyramid level and produces a more target-focused response with less background leakage. This observation is consistent with the quantitative routing-policy comparison in Table~\ref{tab:routing_policy}, where Dynamic Top-2 Routing outperforms Uniform All-Expert Aggregation. We therefore treat the CAM comparison as qualitative support for adaptive expert selection, rather than as standalone proof of a physical scattering mechanism.

\subsubsection{Module-Level CAM Analysis}\label{sec:vis_module}

\begin{figure}[t]
  \centering
  \includegraphics[width=0.96\linewidth]{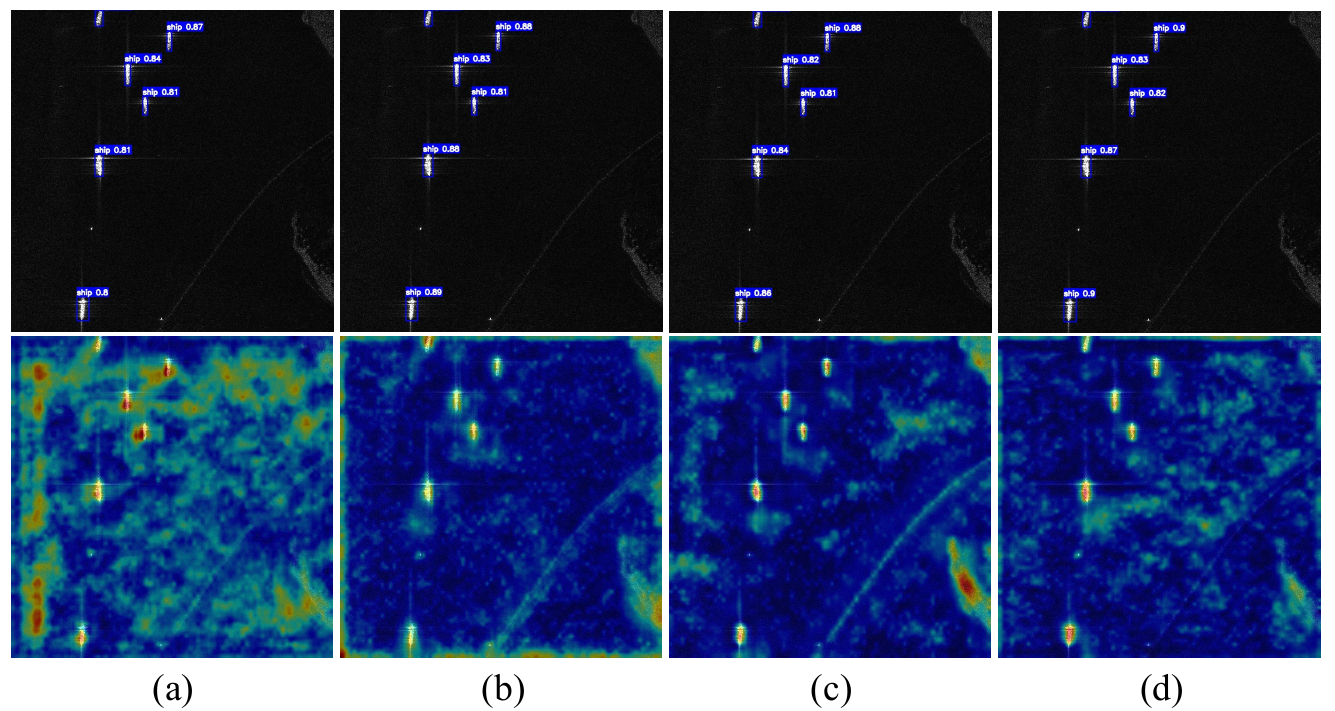}
  \caption{Module-level ablation visualizations on HRSID~\cite{wei2020hrsid} corresponding to Table~\ref{tab:ablation}: (a) DEIM-S baseline; (b) DEIM-S + SARESMoE; (c) DEIM-S + SDEP; and (d) full SARES-DEIM. Each column displays the detection output (top) and corresponding CAM heatmap (bottom). Warmer colors indicate stronger feature responses.}
  \label{fig:vis_module}
\end{figure}

To provide a visual counterpart to the quantitative module-level ablation in Table~\ref{tab:ablation}, we compare detection outputs and CAM heatmaps across four incremental configurations in Fig.~\ref{fig:vis_module}. The four columns correspond exactly to (a) the DEIM-S baseline, (b) the DEIM-S baseline with SARESMoE at P3/P4/P5 but without SDEP, (c) the DEIM-S baseline with SDEP but without SARESMoE, and (d) the full SARES-DEIM model. Each column shows the detection output above its corresponding CAM heatmap.

\noindent\textbf{DEIM-S Baseline.}
The baseline in Fig.~\ref{fig:vis_module}(a) exhibits relatively diffuse CAM responses that extend into the surrounding sea surface and coastal background. Although the ships are detected, several bounding boxes remain comparatively loose around the target structures. In the displayed example, this broader activation pattern is consistent with the baseline's lower strict localization performance in Table~\ref{tab:ablation}.

\noindent\textbf{Baseline + SARESMoE.}
After SARESMoE is inserted at P3/P4/P5 without SDEP, Fig.~\ref{fig:vis_module}(b) shows reduced activation over several non-target background regions and more concentrated responses around the ships. The corresponding boxes are also better aligned with target boundaries in the displayed case. This qualitative change is consistent with SARESMoE improving clutter-aware feature discrimination through input-dependent expert selection.

\noindent\textbf{Baseline + SDEP.}
When SDEP is introduced without SARESMoE, the heatmap in Fig.~\ref{fig:vis_module}(c) retains stronger and more structurally defined target responses. The associated boxes appear tighter than those of the baseline, consistent with SDEP preserving high-resolution spatial cues from the P2 feature. Some surrounding activation remains visible, which is expected because this configuration does not include the expert-routing mechanism for clutter-aware refinement.

\noindent\textbf{Full SARES-DEIM.}
The full model in Fig.~\ref{fig:vis_module}(d) combines the more concentrated background response associated with SARESMoE and the detail-preserving response associated with SDEP. In the displayed example, its heatmap is among the more target-focused of the four configurations, and its predicted boxes closely follow the ship structures. This qualitative result is consistent with the complementary behavior quantified in Table~\ref{tab:ablation}: SARESMoE contributes adaptive clutter-aware refinement, while SDEP preserves fine-grained spatial information. The visualization is used as supporting evidence, whereas the quantitative ablation remains the primary basis for this conclusion.

\FloatBarrier
\raggedbottom
\section{Conclusion}

In this paper, we present \textsc{SARES-DEIM}, a domain-aware detection framework built upon the DETR paradigm and tailored for SAR ship detection. By rethinking the representation bottlenecks in complex SAR environments, we introduce SARESMoE, a SAR-aware Expert Selection Mixture-of-Experts module that dynamically routes features to specialized wavelet, spatial, and frequency-oriented experts for clutter-aware feature refinement. To improve the localization of diminutive targets, we further design the SDEP neck, which utilizes Space-to-Depth convolutions to transmit high-resolution spatial details to the prediction heads.

Evaluations on the HRSID and SAR-Ship-Dataset benchmarks show the effectiveness of \textsc{SARES-DEIM}. On HRSID, our method achieves $mAP_{50:95}$ of 76.4\% and $mAP_{50}$ of 93.8\%, showing competitive performance among the compared YOLO-series, DETR-style, and SAR-specific detectors. The added efficiency analysis further shows that SARES-DEIM should be understood as a capacity-computation trade-off: it stores a larger expert bank while keeping active inference cost at 72.6 GFLOPs, 136.5 FPS under batch size 8, and 37.6 ms batch-size-1 latency on the reported A40 setting. In future work, we plan to extend this domain-aware routing mechanism to multi-modal (SAR-Optical) fusion scenarios and further investigate architectural optimizations to balance representational capacity with computational requirements across diverse deployment platforms.

\bibliographystyle{IEEEtran}
\bibliography{main}

@ARTICLE{wei2020hrsid,
    author  = {Wei, S. and Zeng, X. and Qu, Q. and Wang, M. and Su, H. and Shi, J.},
    title   = {{HRSID}: A high-resolution {SAR} images dataset for ship detection and instance segmentation},
    journal = {IEEE Access},
    volume  = {8},
    year    = {2020},
    pages   = {120234--120254},
    doi     = {10.1109/ACCESS.2020.3005861}
}

@ARTICLE{wang2019sar,
    author  = {Wang, Y. and Wang, C. and Zhang, H. and Dong, Y. and Wei, S.},
    title   = {A {SAR} dataset of ship detection for deep learning under complex backgrounds},
    journal = {Remote Sensing},
    volume  = {11},
    number  = {7},
    year    = {2019},
    pages   = {765},
    doi     = {10.3390/rs11070765}
}

@ARTICLE{zhang2021ssdd,
    author  = {Zhang, Tianwen and Zhang, Xiaoling and Li, Jianwei and Xu, Xiaowo and Wang, Baoyou and Wei, Shunjun},
    title   = {{SAR} Ship Detection Dataset ({SSDD}): Official Release and Comprehensive Data Analysis},
    journal = {Remote Sensing},
    volume  = {13},
    number  = {18},
    year    = {2021},
    pages   = {3690},
    doi     = {10.3390/rs13183690}
}

@misc{lin2014microsoft,
      title={Microsoft COCO: Common Objects in Context}, 
      author={Tsung-Yi Lin and Michael Maire and Serge Belongie and Lubomir Bourdev and Ross Girshick and James Hays and Pietro Perona and Deva Ramanan and C. Lawrence Zitnick and Piotr Dollár},
      year={2015},
      eprint={1405.0312},
      archivePrefix={arXiv},
      primaryClass={cs.CV},
      url={https://arxiv.org/abs/1405.0312}, 
}

@MISC{albanwan2024imagefusionremotesensing,
    author        = {Albanwan, Hessah and Qin, Rongjun and Tang, Yang},
    title         = {Image Fusion in Remote Sensing: An Overview and Meta Analysis},
    year          = {2024},
    eprint        = {2401.08837},
    archivePrefix = {arXiv},
    primaryClass  = {cs.CV},
    url           = {https://arxiv.org/abs/2401.08837}
}

@ARTICLE{jacobs1991moe,
    author  = {Jacobs, R. A. and Jordan, M. I. and Nowlan, S. J. and Hinton, G. E.},
    title   = {Adaptive mixtures of local experts},
    journal = {Neural Computation},
    volume  = {3},
    number  = {1},
    year    = {1991},
    pages   = {79--87},
    doi     = {10.1162/neco.1991.3.1.79}
}

@ARTICLE{shazeer2017moe,
    author  = {Shazeer, N. and Mirhoseini, A. and Maziarz, K. and Davis, A. and Le, Q. and Hinton, G. and Dean, J.},
    title   = {Outrageously large neural networks: The sparsely-gated mixture-of-experts layer},
    journal = {arXiv preprint arXiv:1701.06538},
    year    = {2017}
}

@INPROCEEDINGS{lin2026yolomaster,
    author    = {Lin, Xu and Peng, Jinlong and Gan, Zhenye and Zhu, Jiawen and Liu, Jun},
    title     = {{YOLO-Master}: {MoE}-Accelerated With Specialized Transformers for Enhanced Real-Time Detection},
    booktitle = {Proceedings of the IEEE/CVF Conference on Computer Vision and Pattern Recognition (CVPR)},
    year      = {2026},
    note      = {Accepted}
}

@INPROCEEDINGS{carion2020detr,
    author    = {Carion, Nicolas and Massa, Francisco and Synnaeve, Gabriel and Usunier, Nicolas and Kirillov, Alexander and Zagoruyko, Sergey},
    title     = {End-to-End Object Detection with Transformers},
    booktitle = {Proceedings of the European Conference on Computer Vision (ECCV)},
    year      = {2020},
    pages     = {213--229},
    doi       = {10.1007/978-3-030-58452-8_13}
}

@misc{zhu2020deformabledetr,
    title={Deformable DETR: Deformable Transformers for End-to-End Object Detection}, 
    author={Xizhou Zhu and Weijie Su and Lewei Lu and Bin Li and Xiaogang Wang and Jifeng Dai},
    year={2021},
    eprint={2010.04159},
    archivePrefix={arXiv},
    primaryClass={cs.CV},
    url={https://arxiv.org/abs/2010.04159}, 
}

@INPROCEEDINGS{zhang2023dino,
    author    = {Zhang, Hao and Li, Feng and Liu, Shilong and Zhang, Lei and Su, Hang and Zhu, Jun and Ni, Lionel M. and Shum, Heung-Yeung},
    title     = {{DINO}: {DETR} with Improved DeNoising Anchor Boxes for End-to-End Object Detection},
    booktitle = {Proceedings of the International Conference on Learning Representations (ICLR)},
    year      = {2023},
    url       = {https://openreview.net/forum?id=3mRwyG5one}
}

@INPROCEEDINGS{lv2023detrs,
    author    = {Zhao, Yian and Lv, Wenyu and Xu, Shangliang and Wei, Jinman and Wang, Guanzhong and Dang, Qingqing and Liu, Yi and Chen, Jie},
    title     = {{DETRs} Beat {YOLOs} on Real-Time Object Detection},
    booktitle = {Proceedings of the IEEE/CVF Conference on Computer Vision and Pattern Recognition (CVPR)},
    year      = {2024},
    pages     = {16965--16974},
    doi       = {10.1109/CVPR52733.2024.01605}
}

@INPROCEEDINGS{peng2025dfine,
    author    = {Peng, Yansong and Li, Hebei and Wu, Peixi and Zhang, Yueyi and Sun, Xiaoyan and Wu, Feng},
    title     = {{D-FINE}: Redefine Regression Task of {DETRs} as Fine-Grained Distribution Refinement},
    booktitle = {Proceedings of the International Conference on Learning Representations (ICLR)},
    year      = {2025},
    url       = {https://openreview.net/forum?id=ottpe9H5B1}
}

@INPROCEEDINGS{huang2025deim,
    author    = {Huang, Shihua and Lu, Zhichao and Cun, Xiaodong and Yu, Yongjun and Zhou, Xiao and Shen, Xi},
    title     = {{DEIM}: {DETR} With Improved Matching for Fast Convergence},
    booktitle = {Proceedings of the IEEE/CVF Conference on Computer Vision and Pattern Recognition (CVPR)},
    year      = {2025},
    pages     = {15162--15171}
}

@ARTICLE{yolov8,
    author  = {Yaseen, M.},
    title   = {What is {YOLOv8}: An in-depth exploration of the internal features of the next-generation object detector},
    journal = {arXiv preprint arXiv:2408.15857},
    year    = {2024}
}

@ARTICLE{yolov8_ultralytics,
    author  = {Jocher, G. and Chaurasia, A. and Qiu, J.},
    title   = {Ultralytics {YOLOv8}},
    journal = {GitHub repository},
    year    = {2023},
    note    = {\url{https://github.com/ultralytics/ultralytics}, v8.0.0}
}

@ARTICLE{yolov11,
    author  = {Khanam, R. and Hussain, M.},
    title   = {{YOLOv11}: An overview of the key architectural enhancements},
    journal = {arXiv preprint arXiv:2410.17725},
    year    = {2024}
}

@ARTICLE{yolo11_ultralytics,
    author  = {Jocher, G. and Qiu, J.},
    title   = {Ultralytics {YOLO11}},
    journal = {GitHub repository},
    year    = {2024},
    note    = {\url{https://github.com/ultralytics/ultralytics}, v11.0.0}
}

@ARTICLE{yolov12,
    author  = {Tian, Y. and Ye, Q. and Doermann, D.},
    title   = {{YOLO12}: Attention-centric real-time object detectors},
    journal = {arXiv preprint arXiv:2502.12524},
    year    = {2025}
}

@ARTICLE{FasterRCNN,
    author  = {Lin, Zhao and Ji, Kefeng and Leng, Xiangguang and Kuang, Gangyao},
    title   = {Squeeze and Excitation Rank {Faster R-CNN} for Ship Detection in {SAR} Images}, 
    journal = {IEEE Geoscience and Remote Sensing Letters}, 
    volume  = {16},
    number  = {5},
    year    = {2019},
    pages   = {751--755},
    doi     = {10.1109/LGRS.2018.2882551}
}

@INPROCEEDINGS{SSD,
    author    = {Liu, Wei and Anguelov, Dragomir and Erhan, Dumitru and Szegedy, Christian and Reed, Scott and Fu, Cheng-Yang and Berg, Alexander C.},
    title     = {{SSD}: Single Shot MultiBox Detector},
    booktitle = {Proceedings of the European Conference on Computer Vision (ECCV)},
    year      = {2016},
    pages     = {21--37},
    doi       = {10.1007/978-3-319-46448-0_2}
}

@INPROCEEDINGS{FCOS,
    author    = {Tian, Zhi and Shen, Chunhua and Chen, Hao and He, Tong},
    title     = {{FCOS}: Fully Convolutional One-Stage Object Detection},
    booktitle = {Proceedings of the IEEE/CVF International Conference on Computer Vision (ICCV)},
    year      = {2019},
    pages     = {9626--9635},
    doi       = {10.1109/ICCV.2019.00972}
}

@ARTICLE{CSCF-Net,
    author  = {Qi, Liangang and Huang, Chen and Guo, Qiang},
    title   = {Cross-Scale Context-Aware Ship Detection in {SAR} Images Using {CSCF-Net}}, 
    journal = {IEEE Geoscience and Remote Sensing Letters}, 
    volume  = {23},
    year    = {2026},
    pages   = {1--5},
    doi     = {10.1109/LGRS.2025.3645569}
}

@ARTICLE{SAR-D-FINE,
    author  = {Fan, Xiaobing and Xing, Bowen and Wang, Xingchen and Liu, Hongdan and Yan, Chuanxu and Zhi, Pengfei},
    title   = {{SAR-D-FINE}: A Context-Aware Detector for Small and Densely Packed Ship Detection in {SAR} Imagery}, 
    journal = {IEEE Geoscience and Remote Sensing Letters}, 
    volume  = {23},
    year    = {2026},
    pages   = {1--5},
    doi     = {10.1109/LGRS.2025.3640683}
}

@ARTICLE{OptiSAR-Net,
  author={Dong, Jun and Feng, Jiewen and Tang, Xiaoyu},
  journal={IEEE Transactions on Geoscience and Remote Sensing}, 
  title={OptiSAR-Net: A Cross-Domain Ship Detection Method for Multisource Remote Sensing Data}, 
  year={2024},
  volume={62},
  number={},
  pages={1-11},
  doi={10.1109/TGRS.2024.3502447}
}

@ARTICLE{FINet,
  author={Hu, Qi and Hu, Shaohai and Liu, Shuaiqi and Xu, Shuwen and Zhang, Yu-Dong},
  journal={IEEE Transactions on Geoscience and Remote Sensing}, 
  title={FINet: A Feature Interaction Network for SAR Ship Object-Level and Pixel-Level Detection}, 
  year={2022},
  volume={60},
  number={},
  pages={1-15},
  doi={10.1109/TGRS.2022.3222636}
}

@ARTICLE{CoAM-Darknet,
  author={Yang, Xi and Zhang, Xin and Wang, Nannan and Gao, Xinbo},
  journal={IEEE Transactions on Geoscience and Remote Sensing}, 
  title={A Robust One-Stage Detector for Multiscale Ship Detection With Complex Background in Massive SAR Images}, 
  year={2022},
  volume={60},
  number={},
  pages={1-12},
  doi={10.1109/TGRS.2021.3128060}
}

@INPROCEEDINGS{qiao2022novel,
    author    = {Qiao, Chenchen and Shen, Fei and Wang, Xuejun and Wang, Ruixin and Cao, Fang and Zhao, Sixian and Li, Chang},
    title     = {A Novel Multi-Frequency Coordinated Module for {SAR} Ship Detection},
    booktitle = {Proceedings of the IEEE 34th International Conference on Tools with Artificial Intelligence (ICTAI)},
    year      = {2022},
    pages     = {804--811}
}

@INPROCEEDINGS{weng2023novel,
    author    = {Weng, Weijie and Lin, Weiming and Lin, Feng and Ren, Junchi and Shen, Fei},
    title     = {A novel cross frequency-domain interaction learning for aerial oriented object detection},
    booktitle = {Chinese Conference on Pattern Recognition and Computer Vision (PRCV)},
    year      = {2023},
    pages     = {292--305}
}

@ARTICLE{weng2024enhancing,
    author  = {Weng, Weijie and Wei, Mengwan and Ren, Junchi and Shen, Fei},
    title   = {Enhancing Aerial Object Detection with Selective Frequency Interaction Network},
    journal = {IEEE Transactions on Artificial Intelligence},
    volume  = {1},
    number  = {01},
    year    = {2024},
    pages   = {1--12}
}

@INPROCEEDINGS{li2024lr,
    author    = {Li, Hanqian and Zhang, Ruinan and Pan, Ye and Ren, Junchi and Shen, Fei},
    title     = {{Lr-fpn}: Enhancing remote sensing object detection with location refined feature pyramid network},
    booktitle = {Proceedings of the International Joint Conference on Neural Networks (IJCNN)},
    year      = {2024},
    pages     = {1--8}
}

@INPROCEEDINGS{lin2017fpn,
    author    = {Lin, Tsung-Yi and Doll{\'a}r, Piotr and Girshick, Ross and He, Kaiming and Hariharan, Bharath and Belongie, Serge},
    title     = {Feature Pyramid Networks for Object Detection},
    booktitle = {Proceedings of the IEEE Conference on Computer Vision and Pattern Recognition (CVPR)}, 
    year      = {2017},
    pages     = {936--944},
    doi       = {10.1109/CVPR.2017.106}
}

@ARTICLE{shen2025imagedit,
    author  = {Shen, Fei and Xu, Weihao and Yan, Rui and Zhang, Dong and Shu, Xiangbo and Tang, Jinhui},
    title   = {{IMAGEdit}: Let Any Subject Transform},
    journal = {arXiv preprint arXiv:2510.01186},
    year    = {2025}
}

@ARTICLE{shen2025imagharmony,
    author  = {Shen, Fei and Du, Xiaoyu and Gao, Yutong and Yu, Jian and Cao, Yushe and Lei, Xing and Tang, Jinhui},
    title   = {{IMAGHarmony}: Controllable Image Editing with Consistent Object Quantity and Layout},
    journal = {arXiv preprint arXiv:2506.01949},
    year    = {2025}
}

@ARTICLE{shen2025imaggarment,
    author  = {Shen, Fei and Yu, Jian and Wang, Cong and Jiang, Xin and Du, Xiaoyu and Tang, Jinhui},
    title   = {{IMAGGarment-1}: Fine-Grained Garment Generation for Controllable Fashion Design},
    journal = {arXiv preprint arXiv:2504.13176},
    year    = {2025}
}

@INPROCEEDINGS{shenlong,
    author    = {Shen, Fei and Wang, Cong and Gao, Junyao and Guo, Qin and Dang, Jisheng and Tang, Jinhui and Chua, Tat-Seng},
    title     = {Long-Term {TalkingFace} Generation via Motion-Prior Conditional Diffusion Model},
    booktitle = {Proceedings of the 42nd International Conference on Machine Learning (ICML)},
    year      = {2025}
}

@ARTICLE{shen2024imagpose,
    author  = {Shen, Fei and Tang, Jinhui},
    title   = {{Imagpose}: A unified conditional framework for pose-guided person generation},
    journal = {Advances in Neural Information Processing Systems (NeurIPS)},
    volume  = {37},
    year    = {2024},
    pages   = {6246--6266}
}

@INPROCEEDINGS{shen2025imagdressing,
    author    = {Shen, Fei and Jiang, Xin and He, Xin and Ye, Hu Hollywood and Wang, Cong and Du, Xiaoyu and Li, Zechao and Tang, Jinhui},
    title     = {{Imagdressing-v1}: Customizable virtual dressing},
    booktitle = {Proceedings of the AAAI Conference on Artificial Intelligence},
    volume    = {39},
    number    = {7},
    year      = {2025},
    pages     = {6795--6804}
}

@INPROCEEDINGS{shen2024advancing,
    author    = {Shen, Fei and Ye, Hu and Zhang, Jun and Wang, Cong and Han, Xiao and Wei, Yang},
    title     = {Advancing Pose-Guided Image Synthesis with Progressive Conditional Diffusion Models},
    booktitle = {Proceedings of the International Conference on Learning Representations (ICLR)},
    year      = {2024},
    url       = {https://openreview.net/forum?id=rHzapPnCgT}
}

@InProceedings{sahi2022spdconv,
author="Sunkara, Raja
and Luo, Tie",
editor="Amini, Massih-Reza
and Canu, St{\'e}phane
and Fischer, Asja
and Guns, Tias
and Kralj Novak, Petra
and Tsoumakas, Grigorios",
title="No More Strided Convolutions or Pooling: A New CNN Building Block for Low-Resolution Images and Small Objects",
booktitle="Machine Learning and Knowledge Discovery in Databases",
year="2023",
publisher="Springer Nature Switzerland",
address="Cham",
pages="443--459",
isbn="978-3-031-26409-2"
}

@INPROCEEDINGS{wtconv2024,
    title     = {Wavelet Convolutions for Large Receptive Fields},
    author    = {Finder, Shahaf E and Amoyal, Roy and Treister, Eran and Freifeld, Oren},
    booktitle = {European Conference on Computer Vision},
    year      = {2024},
}

@INPROCEEDINGS{ghostnet2020,
  author={Han, Kai and Wang, Yunhe and Tian, Qi and Guo, Jianyuan and Xu, Chunjing and Xu, Chang},
  booktitle={2020 IEEE/CVF Conference on Computer Vision and Pattern Recognition (CVPR)}, 
  title={GhostNet: More Features From Cheap Operations}, 
  year={2020},
  volume={},
  number={},
  pages={1577-1586},
  doi={10.1109/CVPR42600.2020.00165}}

@INPROCEEDINGS{fadc2024,
  author={Chen, Linwei and Gu, Lin and Zheng, Dezhi and Fu, Ying},
  booktitle={2024 IEEE/CVF Conference on Computer Vision and Pattern Recognition (CVPR)}, 
  title={Frequency-Adaptive Dilated Convolution for Semantic Segmentation}, 
  year={2024},
  volume={},
  number={},
  pages={3414-3425},
  doi={10.1109/CVPR52733.2024.00328}}

@INPROCEEDINGS{fsanet2025,
  author={Zhou, Zhenhuan and He, Along and Wu, Yanlin and Yao, Rui and Xie, Xueshuo and Li, Tao},
  booktitle={2024 IEEE International Conference on Bioinformatics and Biomedicine (BIBM)}, 
  title={Spatial-Frequency Dual Domain Attention Network For Medical Image Segmentation}, 
  year={2024},
  volume={},
  number={},
  pages={4076-4081},
  doi={10.1109/BIBM62325.2024.10822613}
  }

@misc{yuan2025strip,
    author        = {Xinbin Yuan and Zhaohui Zheng and Yuxuan Li and Xialei Liu and Li Liu and Xiang Li and Qibin Hou and Ming-Ming Cheng},
    title         = {Strip R-CNN: Large Strip Convolution for Remote Sensing Object Detection}, 
    year          = {2025},
    eprint        = {2501.03775},
    archivePrefix = {arXiv},
    primaryClass  = {cs.CV},
    url           = {https://arxiv.org/abs/2501.03775}
}

@ARTICLE{11271640,
    author   = {Jiang, Rui and Shi, Hang and Ni, Jiahong and Li, Jiatao and Feng, Yi and Chen, Xinqiang and Li, Yinlin},
    title    = {LSDFormer: Lightweight SAR Ship Detection Enhanced With Efficient Multiattention and Structural Reparameterization}, 
    journal  = {IEEE Journal of Selected Topics in Applied Earth Observations and Remote Sensing}, 
    volume   = {19},
    year     = {2026},
    pages    = {1359--1377},
    doi      = {10.1109/JSTARS.2025.3639164}
}

@ARTICLE{11258897,
    author   = {Cui, Hongbo and Li, Tong and Su, Nan and Yan, Yiming and Feng, Shou and Zhao, Chunhui and He, Jiayue and Gu, Fangning},
    title    = {MSF-SAR: A Multiscale Fusion Method for Small Ship Detection in SAR Images}, 
    journal  = {IEEE Journal of Selected Topics in Applied Earth Observations and Remote Sensing}, 
    volume   = {19},
    year     = {2026},
    pages    = {2200--2212},
    doi      = {10.1109/JSTARS.2025.3634508}
}

@ARTICLE{10468641,
    author   = {Tang, Xiao and Zhang, Jiufeng and Xia, Yunzhi and Xiao, Huanlin},
    title    = {DBW-YOLO: A High-Precision SAR Ship Detection Method for Complex Environments}, 
    journal  = {IEEE Journal of Selected Topics in Applied Earth Observations and Remote Sensing}, 
    volume   = {17},
    year     = {2024},
    pages    = {7029--7039},
    doi      = {10.1109/JSTARS.2024.3376558}
}

@ARTICLE{11016180,
    author   = {Guo, Yue and Chen, Shiqi and Zhan, Ronghui and Wang, Wei and Zhang, Jun},
    title    = {Deformable Feature Fusion and Accurate Anchors Prediction for Lightweight SAR Ship Detector Based on Dynamic Hierarchical Model Pruning}, 
    journal  = {IEEE Journal of Selected Topics in Applied Earth Observations and Remote Sensing}, 
    volume   = {18},
    year     = {2025},
    pages    = {15019--15036},
    doi      = {10.1109/JSTARS.2025.3574184}
}

@ARTICLE{10811768,
    author   = {Fang, Minding and Gu, Yu and Peng, Dongliang},
    title    = {FEVT-SAR: Multicategory Oriented SAR Ship Detection Based on Feature Enhancement Vision Transformer}, 
    journal  = {IEEE Journal of Selected Topics in Applied Earth Observations and Remote Sensing}, 
    volume   = {18},
    year     = {2025},
    pages    = {2704--2717},
    doi      = {10.1109/JSTARS.2024.3520956}
}

@ARTICLE{SUN2026105127,
    author  = {Sun, Zhongzhen and Zhang, Xianghui and Leng, Xiangguang and Wu, Xueqi and Xiong, Boli and Ji, Kefeng and Kuang, Gangyao},
    title   = {{KFIA-Net}: A Knowledge Fusion and Imbalance-Aware Network for Multi-Category {SAR} Ship Detection},
    journal = {International Journal of Applied Earth Observation and Geoinformation},
    volume  = {146},
    year    = {2026},
    pages   = {105127},
    doi     = {10.1016/j.jag.2026.105127}
}

@ARTICLE{10964367,
    author  = {Sun, Zhongzhen and Leng, Xiangguang and Zhang, Xianghui and Zhou, Zheng and Xiong, Boli and Ji, Kefeng and Kuang, Gangyao},
    title   = {Arbitrary-Direction {SAR} Ship Detection Method for Multiscale Imbalance},
    journal = {IEEE Transactions on Geoscience and Remote Sensing},
    volume  = {63},
    year    = {2025},
    pages   = {1--21},
    doi     = {10.1109/TGRS.2025.3559701}
}

@ARTICLE{rs13214209,
    author  = {Sun, Zhongzhen and Leng, Xiangguang and Lei, Yu and Xiong, Boli and Ji, Kefeng and Kuang, Gangyao},
    title   = {{BiFA-YOLO}: A Novel {YOLO}-Based Method for Arbitrary-Oriented Ship Detection in High-Resolution {SAR} Images},
    journal = {Remote Sensing},
    volume  = {13},
    number  = {21},
    year    = {2021},
    pages   = {4209},
    doi     = {10.3390/rs13214209}
}

\makeatletter
\def\@IEEEBIOskipN{2\baselineskip}
\makeatother




\vfill

\end{document}